\documentclass[final]{cvpr}

\usepackage{times}
\usepackage{epsfig}
\usepackage{graphicx}
\usepackage{amsmath}
\usepackage{amssymb}
\usepackage{multirow}
\usepackage{enumitem}
\usepackage{textcomp,booktabs}
\usepackage[dvipsnames]{color}
\usepackage{colortbl}
\usepackage[table]{xcolor}
\definecolor{lightgray}{gray}{0.94}
\definecolor{lightgreen}{rgb}{0.9, 0.98, 0.9}
\definecolor{lightyellow}{rgb}{0.95, 0.99, 0.8}
\definecolor{lightblue}{rgb}{0.9, 0.96, 0.99}
\definecolor{blue}{rgb}{0.0, 0.47, 0.75}
\definecolor{darkgreen}{rgb}{0.0, 0.5, 0.0}
\definecolor{brickred}{rgb}{0.8, 0.25, 0.33}
\usepackage{dsfont}
\usepackage{boldline}
\usepackage{bm}
\usepackage{nopageno}
\usepackage{algorithm}
\usepackage{algorithmic}

\usepackage[pagebackref=false,breaklinks=true,letterpaper=true,colorlinks,bookmarks=false]{hyperref}

\newcommand{\myparagraph}[1]{\vspace{2pt}\noindent{\bf{#1}}}

\definecolor{lightblue}{rgb}{0.9, 0.96, 0.99}
\definecolor{babypink}{rgb}{0.96, 0.88, 0.88}
\definecolor{lightgreen}{rgb}{0.9, 0.98, 0.9}
\newcommand{\up}[1]{\cellcolor{babypink!60}}
\newcommand{\down}[1]{\cellcolor{lightgreen!60}}

\begin{document}

\title{
Distilling Audio-Visual Knowledge by Compositional Contrastive Learning
}

\author{Yanbei Chen\textsuperscript{1}, \hspace{3pt} Yongqin Xian\textsuperscript{2}, \hspace{3pt} A. Sophia Koepke\textsuperscript{1}, \hspace{3pt} Ying Shan\textsuperscript{3}, \hspace{3pt} Zeynep Akata\textsuperscript{1,2,4} \\
{\small
\textsuperscript{1}University of T{\"u}bingen \hspace{2pt}
\textsuperscript{2}MPI for Informatics \hspace{2pt} 
\textsuperscript{3}Tencent PCG \hspace{2pt} 
\textsuperscript{4}MPI for Intelligent Systems} \\
{\small \{yanbei.chen, a-sophia.koepke, zeynep.akata\}@uni-tuebingen.de, yxian@mpi-inf.mpg.de}
}

\maketitle

\begin{abstract}
Having access to multi-modal cues
(e.g. vision and audio) empowers some cognitive tasks to be done faster 
compared to learning from a single modality. 
In this work, we propose to transfer knowledge across heterogeneous modalities, 
even though these data modalities may not be semantically correlated. 
Rather than directly aligning the representations of different modalities, 
we compose audio, image, and video representations across modalities 
to uncover richer multi-modal knowledge. 
Our main idea is to learn 
a compositional embedding that closes the cross-modal semantic gap and captures the task-relevant semantics, 
which facilitates pulling together representations across modalities by compositional contrastive learning.
We establish a new, comprehensive multi-modal distillation benchmark
on three video datasets: UCF101, ActivityNet, and VGGSound. 
Moreover, we 
demonstrate that our model significantly outperforms
a variety of existing knowledge distillation methods 
in transferring audio-visual knowledge to improve  
video representation learning. 
Code is released here: 
\url{https://github.com/yanbeic/CCL}.
\end{abstract}

\vspace{-1em}

\section{Introduction}

Videos often contain informative multi-modal cues, 
such as visual objects, motion, 
auditory events, 
and textual information encoded in   
captions or speech -- 
all of which provide rich, transferrable semantics for representation learning. 
The majority of existing works in video understanding utilise{s} visual-only content for representation learning \cite{wang2011action,simonyan2014two,tran2015learning,wang2016temporal,carreira2017quo,feichtenhofer2019slowfast}. 
Our objective, on the other hand, is to distill the rich {multi-modal knowledge} 
available in networks pre-trained on spatial imagery data 
and temporal auditory data 
to learn more expressive video representations.

\begin{figure}[!t]
\centering
\includegraphics[width=0.48\textwidth]{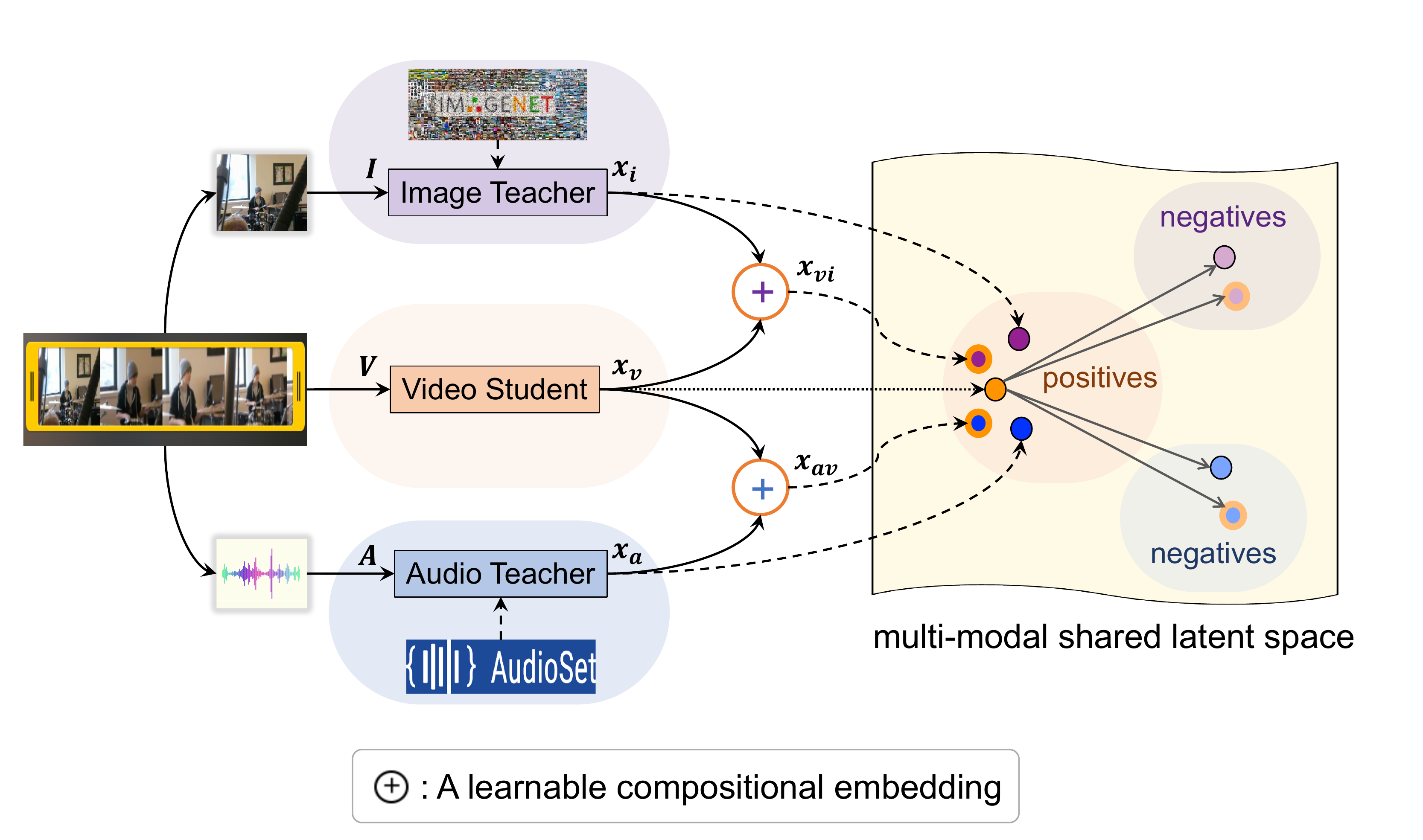}
\caption{Our generic multi-modal distillation framework aligns audio, image, {and} video {representations} in the latent space by compositional contrastive learning, where a compositional embedding is learned to bridge {the} cross-modal semantic gap and capture the task-relevant semantics for more informative knowledge transfer. 
}
\label{fig:illustration}
\vspace{-0.5em}
\end{figure}

In contrast to the standard knowledge distillation techniques \cite{hinton2015distilling,bucilua2006model} 
{which transfer} unimodal knowledge 
learned from the {same} modality and dataset, 
our multi-modal distillation framework uniquely utilises knowledge learned from multiple data modalities. 
Although prior works have considered {cross-modal distillation} \cite{gupta2016cross,aytar2016soundnet,albanie2018emotion,koepke2019visual,afouras2020asr}, 
they generally assume pairwise semantic correspondence 
between two modalities. 
However, in {unconstrained} scenarios, the cross-modal content may not always {be} semantically correlated or temporally aligned, e.g.\ 
a video of {\em applying lipstick}, 
may be accompanied by audio not directly related to the action, 
such as  
{\em music} or {\em speech}.  
On the other hand, similar audio cues, e.g. {\em music}, may accompany 
videos showing distinct visual content, e.g. 
{\em applying lipstick} and {\em playing cello}.

In this work, we tackle a realistic {multi-modal distillation} paradigm 
that can distill {heterogeneous} audio and visual knowledge 
for video representation learning.  
This requires to bridge the cross-modal semantic gap, 
domain gap, as well as dealing with
inconsistent network architectures across modalities.
To address these challenges in a unified formulation, 
we propose 
{compositional contrastive learning} 
-- a novel, generic framework to distill the multi-modal knowledge 
by flexibly plugging in the teacher networks 
pre-trained on different data modalities. 
Specifically, 
{a compositional embedding} is learned to 
close the cross-modal gap 
between the teacher and student networks 
and capture the task-relevant semantics. 
By jointly pulling together the 
teacher, student, and their compositional embeddings
through compositional contrastive learning, 
the multi-modal knowledge  
is transferred to the video student network to 
learn more powerful video representations. 

Our contributions are as follows. 
{\bf (1)}
We propose a novel Compositional Contrastive Learning (CCL) model, featured by 
learnable {compositional embeddings} {that} close the cross-modal semantic gap, and 
a distillation objective which contrasts different modalities jointly in the shared latent space,   
where {class labels} are introduced to distill the multi-modal knowledge in an informative way.
{\bf (2)}
We establish a new benchmark on multi-modal distillation, comparing CCL extensively {to} {\em seven} state-of-the-art distillation methods on {\em three} video datasets: UCF101, ActivityNet, VGGSound, 
in {\em two} tasks: video recognition and video retrieval.
{\bf (3)}
We demonstrate the advantages of our model 
in comparison to the prior state-of-the-art methods, 
and provide an insightful quantitative and qualitative ablative analysis.

\section{Related Work}

\begin{figure*}[!t]
\centering
\includegraphics[width=0.98\textwidth]{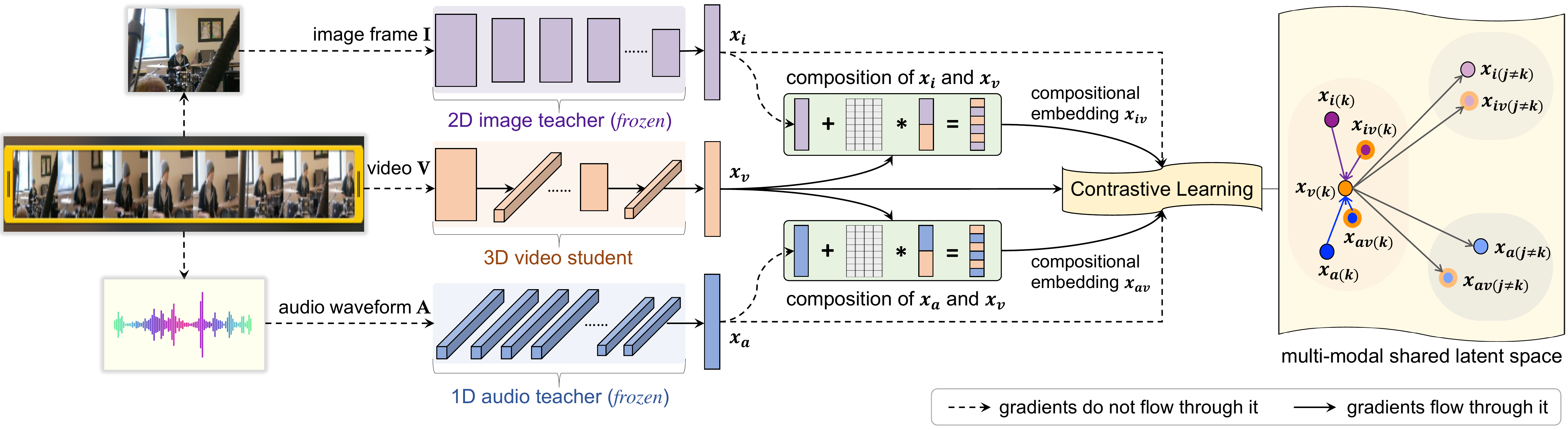}
\caption{ 
Given the image, video, audio  
embeddings  
encoded by individual networks 
(Section \ref{sec:unimodal}),
compositional embeddings are learned to 
{close the cross-modal semantic gap} 
and capture the 
{task-relevant semantics} (Section \ref{sec:crossmodal}), 
which, along with knowledge from the image and audio networks, 
are transferred to the video network 
by compositional contrastive learning in the shared latent space (Section \ref{sec:contrastive}). 
}
\label{fig:model}
\end{figure*}

\myparagraph{Knowledge Distillation.} 
A typical knowledge distillation paradigm follows 
a {teacher-student} learning strategy, 
where the knowledge learned by a large teacher network or 
an ensemble of networks 
is transferred to a {lightweight} student network 
\cite{hinton2015distilling,ba2014deep,bucilua2006model,girdhar2019distinit}. 
In general, supervision signals from the teacher regularise the student network during training, 
as represented by a cross-entropy loss on the {soft targets} \cite{hinton2015distilling},  
or a regression loss on the {pre-softmax activation} \cite{ba2014deep}. 
A line of works reformulate the supervision signals 
for more effective knowledge transfer, 
such as attention transfer \cite{zagoruyko2016paying}, 
probabilistic transfer \cite{passalis2018learning}, 
relation transfer \cite{park2019relational}, 
and correlation transfer \cite{peng2019correlation}. 
Another line of works distill knowledge in a cross-modal context 
\cite{gupta2016cross,aytar2016soundnet,albanie2018emotion,koepke2019visual}, 
such as learning sound representations \cite{aytar2016soundnet},  
or optimising a depth estimation model \cite{gupta2016cross} 
leveraging knowledge from a visual recognition model. 
In contrast to these recent works which assume 
different modalities share similar semantics or physical structures,  
we study a more challenging scenario 
using unconstrained videos with a possible cross-modal semantic gap. 

\myparagraph{Audio-Visual Learning.} 
Audio has been used to assist visual learning or vice versa, e.g. to separate or localise sound in videos 
\cite{owens2018audio,tian2018audio,arandjelovic2018objects,gao2019co}, 
for audio recovery \cite{zhou2019vision}, 
lip reading \cite{afouras2020asr}, 
speech recognition \cite{afouras2018deep}, 
or audio-driven image synthesis \cite{wiles2018x2face,jamaludin2019you}. 
Several self-supervised methods 
have recently been explored for audio-visual learning \cite{owens2016ambient,owens2018learning,alwassel2019self,patrick2020multi}. 
By training the audio and video networks jointly, these works {leverage the semantic correspondence between audio and video} for unsupervised learning on a large video dataset. 
To further scale audio-visual learning to unconstrained audio-video data with possible semantic mismatch, 
we propose to distill the audio-visual knowledge from pre-trained teacher networks 
to regularise the student network. 

\myparagraph{Contrastive Learning.} 
{The} {contrastive loss} was initially proposed 
to learn invariant representations by mapping {similar} inputs to
{nearby} points in {a} latent space \cite{hadsell2006dimensionality}.  
Recently, a family of models 
popularise{d} the idea of contrastive learning for 
self-supervised learning 
\cite{hjelm2018learning,oord2018representation,Federici2020Learning,chen2020improved,chen2020simple}. 
The aim is to maximise the mutual information 
between different views of the same instance \cite{chen2020simple,Federici2020Learning} 
or between the local and global features extracted from the same image \cite{hjelm2018learning}. 
To ensure its success,  
a memory bank is generally used to store a large amount of negative samples,
while various data augmentation techniques are often used to produce 
many views of the same sample. 
Besides self-supervised learning, 
contrastive learning has recently been studied  
in other context{s}, such as knowledge distillation \cite{tian2019contrastive} 
and image generation \cite{park2020contrastive}. 
{To our knowledge}, we {are the first to} introduce contrastive learning 
to distill knowledge across {heterogeneous} modalities. 
Rather than using a vanilla contrastive loss (e.g. {I}nfoNCE \cite{oord2018representation}), 
we formulate a multi-class noise contrastive estimation loss 
that utilises the class labels 
to efficiently associate positives and dissociate negatives across modalities. 

\myparagraph{Video Representations.} 
To represent video information, early works often extract hand-crafted visual features by computing dense trajectories~\cite{wang2011action}, SIFT-3D~\cite{scovanner20073} and HOG-3D~\cite{klaser2008spatio}. Recent advances in video representation learning have been achieved by learning spatiotemporal features with convolutional neural networks (CNNs) on large-scale video datasets, 
as represented by two-stream networks~\cite{simonyan2014two,feichtenhofer2019slowfast}, 3D-CNNs~\cite{tran2015learning, carreira2017quo, hara2018can}  
and factorised 3D-CNNs~\cite{qiu2017learning,tran2018closer}. 
While these approaches are trained using annotated videos, some self-supervised methods learn video representations from a large collection of unlabelled videos by sorting video frames~\cite{xu2019self,kim2019self}, using temporal cycle consistency~\cite{dwibedi2019temporal,wang2019learning} or video colorisation~\cite{vondrick2016anticipating}. Recently, there has been a growing interest in 
learning multi-modal video representations with  
audio signals~\cite{wang2020makes,liu2019use} or video captions~\cite{gabeur2020multi}, 
which both aim to integrate the multi-modal cues 
into a single feature encoding to represent each video. 
Although our model also exploits the multi-modal semantics in videos, 
it uniquely leverages the privileged knowledge from  
multiple modalities without training multiple networks jointly, and processes
{\em only} the video modality at test time for a higher efficiency. 

\section{Compositional Contrastive Learning (CCL)}

Our goal is to distill audio-visual knowledge learnt from 
heterogeneous {audio} and {image} modalities for {video} representation learning, 
while the cross-modal content may be semantically unrelated.  
The image and audio embeddings are 
first extracted from pre-trained teacher networks (Figure \ref{fig:model}, top and bottom), 
and then composed with the video embeddings (Figure \ref{fig:model}, middle) 
to bridge the cross-modal semantic gap 
(Section \ref{sec:crossmodal}). 
The unimodal audio and image embeddings, along with the compositional embeddings, 
are then jointly aligned with the video embeddings 
by compositional contrastive learning 
to distill audio-visual knowledge  
(Section \ref{sec:contrastive}). 
At test time, {\em only} the video student network is deployed for {the} video recognition {or} video retrieval task{s}.  

\subsection{Unimodal Representations of Audio and Vision}
\label{sec:unimodal} 

Given a dataset of $N$ videos 
$\mathcal{D} = \{\textbf{V}_i, y_i\}_{i=1}^{N}$ --
each video belongs to one of $\mathcal{K}$ video categories 
and contains a set of images $\{{\textbf{I}_{ij}}\}_{j=1}^{M_i}$ 
with an audio recording $\textbf{A}_i$, 
we extract the unimodal embeddings via CNNs. 
As audio, image, and video data exhibit  
{heterogeneous} characteristics, 
different network architectures are adopted 
to model the {temporal}, 
{spatial}, or {spatiotemporal} information, as detailed next. 

\myparagraph{Audio Teacher Network.} 
Although the audio and visual content in a video may not be  
semantically related, 
the audio knowledge encodes temporal context 
that offers rich {privileged information} \cite{vapnik2015learning}.  
Given the audio recording ${\bf A}_i$ of the video ${\bf V}_i$, 
the log-mel spectrogram is extracted and 
passed through a pre-trained 1D-CNN 
to obtain an audio embedding, formally referred {to} as $x_a=\theta_{\text{1D-CNN}}({\bf A}_i)$, 
where $x_a$ is a $K_a$-dimensional audio teacher embedding, 
$\theta_{\text{1D-CNN}}$ is the {\em audio teacher network} 
parameterised {by} 1D convolutions to capture the temporal acoustic context. 
 
\myparagraph{Image Teacher Network.} 
The image teacher network is a standard 2D-CNN 
to encode the spatial visual information.  
Given an image frame ${\bf I}_{ij}$ randomly sampled from the video ${\bf V}_i$, 
an image embedding is extracted using the 2D-CNN, 
referred to as $x_{i}=\theta_{\text{2D-CNN}}({\bf I}_{ij})$, 
where $x_{i}$ is a $K_i$-dimensional image teacher embedding, and 
$\theta_{\text{2D-CNN}}$ is the {\em image teacher network}. 
As each video clip contains a set of image frames, 
only one image frame is randomly selected at a time to represent 
the spatial visual content. 
 
\myparagraph{Video Student Network.} 
To distill audio-visual knowledge from two teacher networks, 
a 3D-CNN network customised for video recognition 
is employed to learn the video representations from scratch, 
while mimicking the teacher networks (detailed in Section \ref{sec:contrastive}). 
The video network contains a stack of residual blocks with (2+1)D convolutions, 
which alternates between 2D spatial convolutions
and 1D temporal convolutions
to encode the spatiotemporal visual content. 
Given a $T {\times} H {\times} W {\times} 3$ volume 
of an RGB video clip from video ${\bf V}_i$, 
a video embedding is parameterised by the 3D-CNN, 
formally referred to as 
$x_v=\theta_{\text{3D-CNN}}({\bf V}_i)$, 
where 
$x_v$ is a $K_v$-dimensional video embedding, 
and 
$\theta_{\text{3D-CNN}}$ is the 
{\em video student network} trained by a cross-entropy loss 
to predict a probability distribution $P_v$ over $\mathcal{K}$ video classes: 
\begin{align*}
\mathcal{L}_{ce}^v = \mathcal{L}_{ce}(x_v,k)
 ={-}\log \Big(P_v(k|{x}_{v};\theta_{cls})\Big), 
 \label{eq:ce}
 \end{align*} 
where $k$ is the class label of ${\bf V}_i$, 
$\theta_{cls}$ is the video classifier. 

\subsection{Compositional Multi-Modal Representations} 
\label{sec:crossmodal} 

As aforementioned, 
the student and teacher embeddings 
may be semantically unaligned -- 
an image frame may capture {only partial visual cues} not directly related to the video event, 
while the accompanied audio of
an action video may be irrelevant music or speech. 
To bridge the possible {semantic gap} and {domain gap} across modalities, 
we propose to {rectify} the audio and image teacher embeddings 
by composing the teacher and student embeddings 
and constraining the compositional embeddings with our task objective 
to close the possible semantic gap. 
As the network architectures are nonuniform across different modalities, 
the cross-modal composition is derived at the penultimate layer. 

Formally, the teacher embeddings $x_a$, $x_{i}$ are composed 
with the student embedding $x_v$ 
by learning a residual {on top of} the teacher embeddings. We rectify the teacher embeddings using the following composition function $\mathcal{F}(\cdot,\cdot)$, which learns a residual $f_{\theta}(\cdot,\cdot)$ that fuses two modalities by normalisation, concatenation and a linear projection:  
\begin{equation}
\begin{aligned}
\mathcal{F}_{av}(x_a, x_v) & = x_{av} = x_a + f_{{\theta_{av}}}(x_a, x_v),
\\
\mathcal{F}_{iv}(x_i, x_v) & = x_{iv} = x_i + f_{{\theta_{iv}}}(x_i, x_v),
\end{aligned}
\label{eq:compose}
\end{equation}
where $x_{av}, x_{iv}$ are the compositional embeddings. 
This operation is related to prior works 
that compose multi-modal features \cite{vo2019composing,chen2020image,chen2020learning},  
but ours aims at 
{shifting} the teacher embedding with a learnable {residual}. 
More importantly, 
to constrain the class assignment of the compositional embeddings, 
$\mathcal{F}(\cdot,\cdot)$ is 
optimised by the video classification loss 
(i.e. $\mathcal{L}_\text{ce}$), 
which ensures $x_{av}, x_{iv}$ are assigned to the same video class label as $x_v$. 
The overall classification loss is: 
\begin{align}
    \mathcal{L}_{cls} =  \mathcal{L}_{ce}^v(x_v,k) +  \mathcal{L}_{ce}^{av}(x_{av},k) +  \mathcal{L}_{ce}^{iv}(x_{iv},k), 
\label{eq:cls}
\end{align}
where $\mathcal{L}_{ce}^{av}, \mathcal{L}_{ce}^{iv}$ are imposed on 
the composition functions $\mathcal{F}_{av}(\cdot,\cdot), \mathcal{F}_{iv}(\cdot,\cdot)$. 
In the presence of a cross-modal semantic gap (e.g., the audio and video embeddings $x_a$, $x_v$ belong to different classes), the compositional embeddings are enforced to share the same class label $k$ as $x_v$. 
In other words, the compositional embeddings are learned to 
close the possible semantic gap and 
capture the task-relevant semantics 
to facilitate more informative knowledge transfer.

\subsection{Distilling Audio-Visual Knowledge} 
\label{sec:contrastive} 

Many prior unimodal methods distill knowledge merely in the prediction space by enforcing the student network to output similar predictions as the teacher network 
\cite{hinton2015distilling,ba2014deep,bucilua2006model}. 
However, this strategy cannot be directly applied to multi-modal distillation, given the teacher networks are often pre-trained on heterogeneous task objectives to predict different classes. Thus, we propose to perform contrastive learning in the latent feature space, followed by contrasting the class distributions in the prediction space. 

Given the unimodal embeddings 
and the compositional embeddings, 
we propose to distill the knowledge by 
{pulling} together the positive pairs 
while {pushing} away the negative pairs across modalities.  
The positive pairs could include the images, audios, and videos from the same video class $k$. 
Specifically, for a triplet of audio, video, and their compositional embeddings 
extracted from the video $\textbf{V}_i$, 
the contrastive loss can be formed in 
every pair among them to reinforce their correspondence 
in the shared embedding space. 
Formally, a contrastive loss $\mathcal{L}_{ct}$
(based on InfoNCE \cite{oord2018representation})  
between a pair of audio and video embeddings $x_{v(i)}, x_{a(i)}$ 
can be derived as below. 
\begin{equation}
\hspace{-1em}
\begin{aligned}
\mathcal{L}_{ct}
{=}
{-}\text{log}\frac{\text{exp}(\Phi(x_{v(i)},x_{a(i)})/\tau)}
{\sum_{j=1}^{B}\text{exp}(\Phi(x_{v(j)},x_{a(j)}))/\tau)} 
{=}{-}\text{log} \ p_{av(i)},
\end{aligned}
\label{eq:infonce}
\end{equation}
where $\Phi$ is a cosine similarity scoring function, $\tau$ is the temperature, 
$p_{av(i)}$ is the probability of 
assigning the video embedding $x_{v(i)}$ to its paired audio embedding $x_{a(i)}$ 
against the whole mini-batch of audio embeddings $\{x_{a(j)}\}_{j=1}^B$. 

Although the contrastive loss $\mathcal{L}_{ct}$ (Eq. \eqref{eq:infonce}) has shown its 
success as an {instance-level} self-supervised signal \cite{chen2020simple,chen2020improved}, 
it is not directly applicable in our context,  
given there may exist multiple positive video-audio pairs 
sampled from an identical video class $k$ per batch. 
Therefore, we formulate a multi-class noise contrastive estimation (NCE) loss 
that brings the {class label} $k$ into the loss formulation: 
\begin{equation}
\hspace{-1em}
\begin{aligned}
\mathcal{L}_{nce}(x_v, x_a) 
{=}{-}\frac{1}{B_p}\sum_{j=k}\text{log} \ p_{av(j)}
{-}\frac{1}{B_n}\sum_{j{\neq}k}\text{log}(1 {-} p_{av(j)}),
\end{aligned}
\label{eq:nce}
\end{equation}
where $B_p, B_n$ are the number of positive pairs (from class $k$) 
and negative pairs (not from class $k$) for the video embedding $x_v$ (labelled as $k$). 
When $B_p{=}B_n{=}1$, Eq. \eqref{eq:nce} is equivalent to the vanilla NCE \cite{gutmann2010noise}. 
Note that Eq. \eqref{eq:infonce} 
considers each instance as a class; 
thus, using Eq. \eqref{eq:infonce} would ignore the 
class-level discrimination and blindly treat some positives as false negatives. 
In contrast, our multi-class NCE (Eq. \eqref{eq:nce}) encourages the network to assign higher probabilities to the positives and lower probabilities to the negatives. 

The same multi-class NCE loss can be imposed between the 
video embedding $x_v$ and the compositional embedding $x_{av}$ 
to collectively distill unimodal audio knowledge and 
multi-modal knowledge in 
a {\em compositional} manner: 
\begin{equation}
\begin{aligned}
\mathcal{L}_{a}(x_v, x_a, x_{av}){=}
\lambda\mathcal{L}_{nce}(x_v, x_a)+(1-\lambda)\mathcal{L}_{nce}(x_v, x_{av}),
\end{aligned}
\label{eq:audio}
\end{equation}
where $\lambda$ is a hyperparameter to balance the two terms, 
$\mathcal{L}_{a}$ is the objective from the audio modality. 
Similarly, the above objective can be rewritten by utilising the image modality: 
\begin{equation}
\begin{aligned}
\mathcal{L}_{i}(x_v, x_i, x_{iv}){=} 
\lambda\mathcal{L}_{nce}(x_v, x_i)+(1-\lambda)\mathcal{L}_{nce}(x_v, x_{iv}),
\end{aligned}
\label{eq:image}
\end{equation} 
where $x_i$, $x_{iv}$ are the image embedding and 
the compositional embedding respectively. 
In essence, 
the loss function (Eq.\eqref{eq:audio} or Eq.\eqref{eq:image}) 
serves as a {similarity constraint} to 
align embeddings 
in the multi-modal latent {\em feature space}. 
Since the compositional embeddings are constrained by 
the classifiers for video classification, 
a {similarity constraint} 
can also be imposed  
to further align the {class distributions} in the {\em prediction space}. 
Formally, given the predictive distributions from 
the video network $P_v$ and the composition functions 
$P_{av}$, $P_{iv}$, 
we introduce
the Jensen–Shannon divergence (JSD) \cite{lin1991divergence}
to align $P_v$ with respect to
$P_{av}, P_{iv}$: 
\begin{equation}
\begin{aligned}
\mathcal{L}_{JSD}{=} 
JSD(P_v||P_{av})+JSD(P_v||P_{iv}),
\end{aligned}
\label{eq:jsd}
\end{equation}
where $JSD(P||Q)$ is a symmetric {similarity} measure  
between two distributions, i.e. 
$JSD(P||Q) = (KL(P||Q)+KL(Q||P))/2$.  
Minimising $\mathcal{L}_{JSD}$ contrasts the {class semantics} 
between $P_v$ and $P_{av}, P_{iv}$ in the {prediction space}, 
which is orthogonal to feature-level contrastive learning. 
 
\myparagraph{Learning Objective.} 
The final compositional contrastive learning objective (CCL) 
for distilling audio-visual knowledge in the video recognition task 
can be written as below. 
\begin{equation}
\begin{aligned}
\mathcal{L}_{CCL}=\mathcal{L}_{\text{distill}} + \lambda_{cls}\mathcal{L}_{cls}, 
\text{with} \ \mathcal{L}_{\text{distill}}{=}\mathcal{L}_{i}{+}\mathcal{L}_{a}{+}\mathcal{L}_{JSD}
\end{aligned}
\label{eq:ccl}
\end{equation}
where the distillation objective $\mathcal{L}_{\text{distill}}$ 
optimises the model along with the classification objective $\mathcal{L}_{cls}$ (Eq. \eqref{eq:cls}). 
According to the availability of pre-trained networks from the image and audio modalities, 
$\mathcal{L}_{\text{distill}}$ can be flexibly rewritten as 
$\mathcal{L}_{i}{+}JSD(P_v||P_{iv})$ 
to distill visual-only knowledge, 
or 
$\mathcal{L}_{a}{+}JSD(P_v||P_{av})$ 
to distill audio-only knowledge. 
In our experiments (Section \ref{sec:experiments}),  
we consider audio, visual, and audio-visual distillation 
to study the individual impact of each modality, 
as well as their combinative impact.

\section{Experiments}
\label{sec:experiments}

\myparagraph{Datasets.} 
To establish a comprehensive multi-modal distillation benchmark, 
we use 
the following video datasets. 
{\bf (1) UCF51}~\cite{soomro2012ucf101} is a subset of UCF101 that contain audios in videos, including 6,845 videos from 51 {action} classes, such as {\em baby crawling} and {\em apply lipstick}. We use the public split 1 for evaluation. {\bf (2) ActivityNet}~\cite{caba2015activitynet} contains 14,950 videos, covering a wide range of 200 complex {human daily activities}, such as {\em arm wrestling} and {\em having an ice cream}. We use the default split (10,024 training vs 4,926 validation videos). {\bf (3) VGGSound}~\cite{vedaldi2020vggsound} is a large-scale dataset of 309 audio-visual correspondent event classes from 199,196 videos, such as {\em playing violin} and {\em thunder}. We use 183,730 videos for training and the rest for testing. 
Note that except VGGSound, audio and video in the other two datasets are {\em not} always semantically correlated. 

\myparagraph{Implementation Details.} 
We use 
R(2+1)D-18 \cite{tran2018closer} as the video student network.  
The audio and image teacher networks are the 
1D-CNN14 \cite{kong2019panns} and 2D-ResNet34 \cite{he2016deep}, 
pre-trained on the AudioSet \cite{gemmeke2017audio} and ImageNet \cite{deng2009imagenet}. 
The model weights of {the} teacher networks are {\em kept frozen} during training.  
The video network is trained by SGD 
with a learning rate of $0.001$, a weight decay of $0.0005$.
The batch size is $16$ on UCF51, $64$ on ActivityNet, $256$ on VGGSound. 
The temperature $\tau$ {is} $0.1$, $0.5$ on the image and audio modality. 
The hyperparameters $\lambda, \lambda_{cls}$ are set to 0.5, 1. 
The dimension of the latent feature space is 512. 
As the feature dimensions of all networks are 512, 
we do not add projections upon the networks, 
but linear projections can be added to map all embeddings to the same dimensionality.  
The video clips are cropped to $112{\times}112$. Each clip contains 16 frames. 
We train the model without accessing the AudioSet or ImageNet. 
At test time, {\em only} the video network is used.  
More algorithmic details are given in the {\em supplementary}.

\myparagraph{Tasks and Metrics.} 
We evaluate the quality of video representations 
on video recognition and 
video retrieval tasks. 
The video network is trained for recognition, 
and tested for both recognition and retrieval. 
For recognition, top-1 accuracy (\%) is reported
to show the classification accuracy on {the} test set. 
For retrieval, R@K (recall@K, \%) is reported, i.e. the top $k$ nearest neighbours (kNN) contain videos from the same class 
as the query videos. 
In kNN retrieval \cite{buchler2018improving}, 
videos in the test set are used as queries 
and videos in the training set are the retrieval targets.
For each video, multiple clip-level features 
are extracted by applying a sliding window. 
The feature {per video} is computed 
by averaging all the clip-level features. 
A cosine distance metric is finally adopted to 
measure the pairwise similarity in kNN retrieval. 

{
\setlength{\tabcolsep}{7pt}
\renewcommand{\arraystretch}{1.1}
\begin{table}[!t]
	\centering
	\resizebox{\columnwidth}{!}{
	\begin{tabular}{l|ccc|ccc}	
	    \multirow{2}{5pt}{Method} 
		& \multicolumn{3}{c|}{UCF51} 
		& \multicolumn{3}{c}{ActivityNet} \\ 
		& \bf A & \bf I & \bf AI 
		& \bf A & \bf I & \bf AI \\ \hline
		baseline & 
		57.5 & 57.5 & 57.5 & 
		32.6 & 32.6 & 32.6\\ \hline
		FitNet & 
		48.4  & 67.4  & 62.4  & 
		21.3  & 45.8  & 34.6   \\
		PKT & 
		53.2  & 58.2  & 62.0  & 
		33.4  & 35.4  & 35.1  \\
		COR & 
		57.7  & 65.5  & 66.3  & 
		31.4  & 43.1  & 41.7  \\
		RKD & 
		53.0  & 55.4  & 58.2  & 
		-  & 34.3  & -  \\
		CRD & 
		60.3  & 61.4  & 63.2  & 
		36.4  & 37.3  & 36.6  \\
		IFD & 
		56.3  & 54.2  & 64.2  & 
		34.6  & 33.8  & 35.4  \\
		CMC & 
		59.2  & 60.4  & 63.1  & 
		34.4  & 23.7  & 33.9  \\
		\hline 
		\bf CCL & 
		\bf 64.9  & \bf  69.1  & \bf 70.0  & 
		\bf 36.5  & \bf 46.3  & \bf 47.3   
	\end{tabular}
	}
    	\vskip 0.1em
	\caption{Video recognition on UCF51 and ActivityNet. Metric: Top1 accuracy (\%). 
	Knowledge is transferred from A: audio; 
	I: image; 
	or AI: audio and visual modalities
	to improve the video recognition model 
	(`-': the model is not converged). 
	} 
	\label{tab:recognition}
\end{table}
}

\subsection{Comparing with the State of the Art}
\label{sec:result}

\myparagraph{Baseline and State-of-the-art Models.} 
For a fair and comprehensive evaluation, 
we propose a multi-modal distillation benchmark,  
comparing our CCL to  
a simple {\bf baseline} model without distillation and 
{\em seven} state-of-the-art distillation methods.  
We train each model similar to CCL, but replace the distillation objective 
based on their open-source implementation. 
In the following, we describe their distillation objectives in brief. 
{\bf FitNet} \cite{romero2014fitnets} 
aligns the representations of the teacher and student by regression. 
{\bf PKT} \cite{passalis2018learning} 
models the feature distribution by a probabilistic model, and
matches the distributions between the student and teacher by a divergence metric. 
{\bf CCKD} \cite{peng2019correlation} 
transfers the correlation among instances in {the} feature space from the teacher to the student by regression. 
{\bf RKD} \cite{park2019relational} 
transfers the distance-wise and angle-wise relations of features from the teacher to {the} student by penalising differences in relations. 
{\bf CRD} \cite{tian2019contrastive} 
transfers knowledge by instance-level contrastive learning and uses a large memory bank to store negative samples. 
{\bf IFD} \cite{passalis2020heterogeneous} 
trains the student network to mimic the teacher's information flow derived by a probabilistic model. 
{\bf CMC} \cite{tian2020contrastive} 
is a cross-view learning method 
to align different views of the same instances 
by contrastive learning.

\noindent {\em Remark.} 
The above methods are proposed 
based on an assumption that the teacher and student are trained on
the unimodal data or {on the} same task objective; 
while we uniquely {consider} to distill knowledge learned from {heterogeneous} multiple data modalities. 
Our model also differs in several aspects compared to other contrastive learning methods (CRD, CMC). 
First, we introduce learnable {compositional embeddings} 
to close {the} cross-modal semantic gap and capture task-relevant semantics.  
Second, rather than treating each instance as one class, 
our objective exploits the {class labels} to enhance discrimination of different classes. 
Third, we do not use a large memory bank, 
thus greatly lowering the computation cost to derive the contrastive loss.

{
\setlength{\tabcolsep}{5pt}
\renewcommand{\arraystretch}{1.1}
\begin{table*}[t]
    \centering
	\resizebox{\linewidth}{!}{%
	\begin{tabular}{l|ccc|ccc|ccc|ccc|ccc|ccc}		
	         \multirow{2}{5pt}{Setup} 
		& \multicolumn{9}{c|}{UCF51} 
		& \multicolumn{9}{c}{ActivityNet} \\ 
		& \multicolumn{3}{c|}{ \bf A} & \multicolumn{3}{c|}{ \bf I} & \multicolumn{3}{c|}{ \bf AI}
		& \multicolumn{3}{c|}{ \bf A} & \multicolumn{3}{c|}{ \bf I} & \multicolumn{3}{c}{ \bf AI} \\ 
		Metrics & 
		R1 & R5 & R10 & R1 & R5 & R10 & R1 & R5 & R10 & 
		R1 & R5 & R10 & R1 & R5 & R10 & R1 & R5 & R10  \\ \hline
		baseline & 
		57.3 & 65.3 & 68.9 &
		57.3 & 65.3 & 68.9 &
		57.3 & 65.3 & 68.9 &
		29.0 & 46.5 & 54.9 & 
		29.0 & 46.5 & 54.9 & 
		29.0 & 46.5 & 54.9 
		\\ \hline
		FitNet & 
		31.9 & 42.5 & 47.6 & 
		51.4 & 66.9 & 73.7 & 
		61.2 & 68.7 & 72.4 &
		16.5 & 32.8 & 42.1 & 
		30.7 &  52.6 &  62.5 & 
		30.5 & 48.3 & 57.0 
		\\ 
		PKT & 
		48.4 & 57.2 & 61.8 & 
		53.0 & 63.4 & 68.9 & 
		61.2 & 69.1 & 72.2 &
		26.4 & 44.4 & 53.3 & 
		28.1 & 48.1 & 57.4 & 
		30.3 & 47.6 & 56.2 
		\\
		COR & 
		51.7 & 58.6 & 61.4 & 
		56.1 & 66.9 & 73.7 & 
		52.8 & 65.1 & 72.6 &
		27.3 & 45.7 & 54.9 & 
		33.2 & 55.2 & 64.6 & 
		30.2 & 51.7 & 61.7 
		\\
		RKD & 
		46.6 & 56.8 & 61.8 & 
		51.2 & 59.7 &  65.3 & 
		55.8 & 63.7 & 66.9 & 	
		- & - & - &
		27.1 & 47.1 & 56.1 &
		- & - & - \\
		CRD & 
		59.5 & 65.8 & 68.0 & 
		59.1 & 66.5 & 69.1 & 
		61.0 &  66.9 & 70.1 &
		\bf 31.6 & \bf 50.4 & \bf 58.9 & 
		32.3 & 50.7 & 58.6 & 
		33.1 & 49.8 & 58.2 
		\\
		IFD & 
		53.8 & 60.8 & 65.6 & 
		48.0 & 58.6 & 65.3 & 
		64.3 & 71.1 & 74.2 &
		28.9 & 47.1 & 56.1 & 
		25.3 & 44.4 & 54.4 & 
		30.0 & 47.6 & 56.1 
		\\
		CMC & 
		57.9 & 64.5 & 67.7 & 
		60.1 & 63.5 & 65.0 & 
		62.9 & 69.0 & 71.7 &
		30.0 & 49.1 & 57.6 & 
		25.2 & 44.2 & 53.6 & 
		30.9 & 48.2 & 55.9 
		\\
		\hline 
		\bf CCL & 
		\bf 62.9 & \bf 68.0 & \bf 70.4 &
		\bf 66.8 & \bf 73.5  & \bf 76.5 & 
		\bf 67.6 & \bf 72.3 & \bf 74.7 &
		30.6 & 49.1 & 57.3 & \bf 38.1 & \bf 58.8 & \bf 67.4 & \bf 39.5 & \bf 59.3 & \bf 67.4 
		\\
	\end{tabular}
	}
    \vskip 0.1em
	\caption{Video retrieval on UCF51 and ActivityNet. 
	Metrics: R@K ($K=1,5,10$, \%).  
	Knowledge is transferred from A: audio; 
	I: image; 
	or AI: audio and visual modalities
	to improve the video recognition model ( 
	`-': the model is not converged). 
	} 
	\label{tab:retrieval}
	\vskip -0.5em
\end{table*}
}

\begin{figure}[t]
\centering
\includegraphics[width=0.47\textwidth]{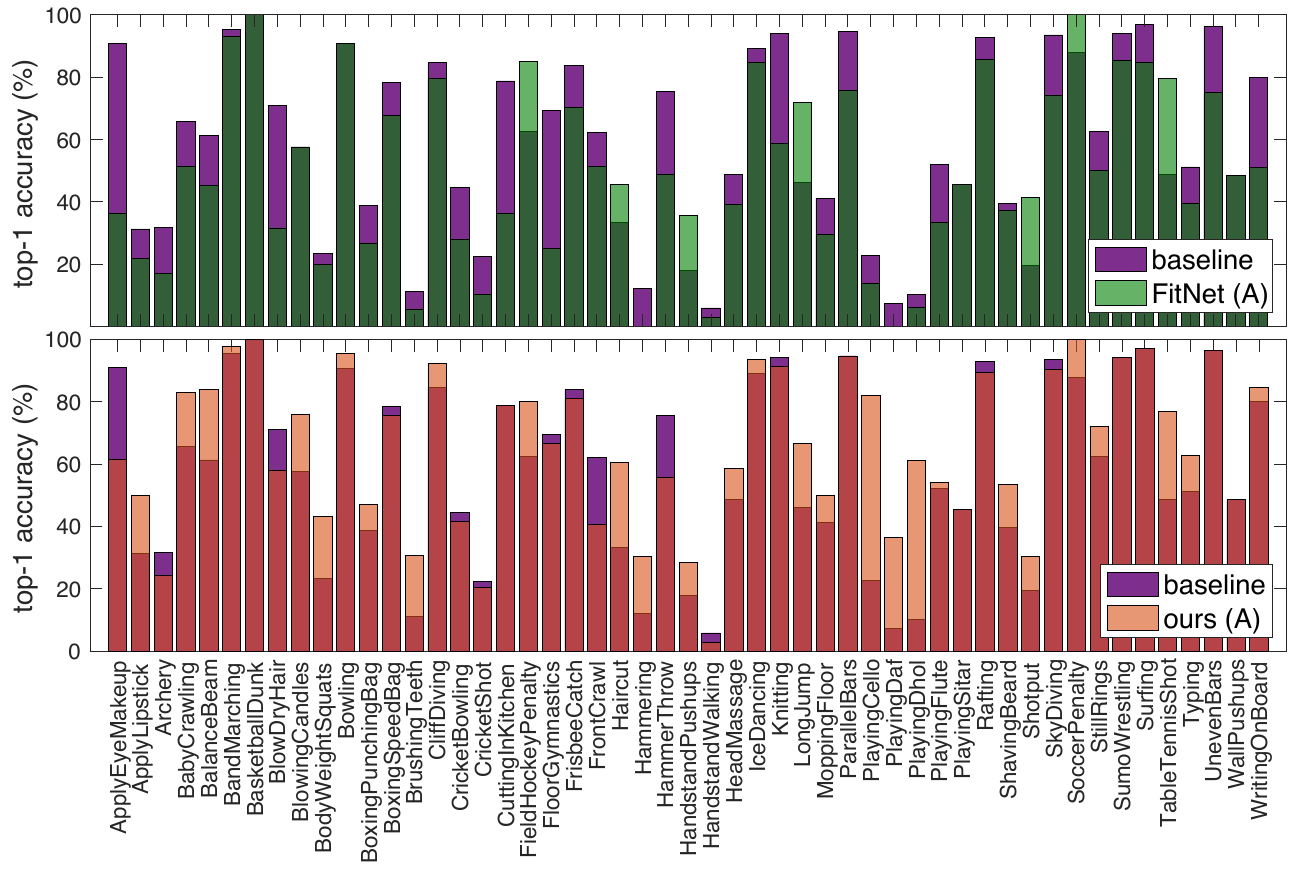}
\caption{
The per-class top-1 video recognition accuracy for audio distillation on UCF51 comparing FitNet and CCL wrt the baseline. Light green and orange: FitNet and CCL outperform the baseline. 
}
\label{fig:perclass}
\vskip -0.5em
\end{figure}

\myparagraph{Video Recognition.} 
In Table \ref{tab:recognition}, 
we evaluate on three setups: 
audio distillation (A), 
visual distillation (I), 
and 
audio-visual distillation (AI) on {the} UCF51 and ActivityNet datasets. 
As shown in Table \ref{tab:recognition}, distilling knowledge from all modalities, our CCL obtains the state-of-the-art consistently and improves over the prior methods. Specifically, 
when it comes to visual distillation with {the} image modality (I), 
we observe that FitNet is a strong competitor on both datasets. Similarly, when it comes to audio distillation (A), CRD obtains the prior state-of-the-art on both datasets. However, this observation does not hold on the other modality for both FitNet and CRD. 
On the other hand, our CCL makes better use of the knowledge learned from either the image or audio modality and thus outperforms all methods on both datasets, e.g. significantly boosting the baseline by 7.4\% (64.9-57.5), 11.5\% (69.1-57.6) on UCF51 (A) and (I). 

On audio-visual distillation (AI), our CCL significantly outperforms the prior state-of-the-art on both datasets, obtaining an impressive result of 70.0\% (vs 66.3\% by COR) on UCF51 and 47.3\% (vs 41.7\% by COR) on ActivityNet. Although the prior state-of-the-art may perform well on audio or visual distillation (e.g. CRD, FitNet), this behaviour does not remain when it comes to multi-modal distillation, i.e. when using audio and visual modalities jointly. The new state-of-the-art obtained by our CCL in the setup of (AI) indicates its capability to distill the complementary knowledge from heterogeneous modalities in a robust manner.

\myparagraph{Closer Look at Audio Distillation in Video Recognition.} 
Although FitNet is a strong competitor in visual distillation, it does not perform as well in audio distillation. To better understand this discrepancy, we closely inspect the per-class top-1 video recognition accuracy on UCF51 using FitNet and our CCL in audio distillation. 
Our results in Figure \ref{fig:perclass} indicate that  
our model outperforms the baseline in most classes, 
while the performance of FitNet degrades in most classes (44 out of 51). 
As audio and video content are generally not semantically related on UCF51,  
when we look at individual classes, 
we find {that} FitNet mostly predicts incorrectly when the audio is not highly in line with the video content, e.g. in {the} class {\em writing on board} where FitNet fails, most videos show a person speaking while writing on board, and the audio is weakly related to the action. 
In {the} class {\em table-tennis shot} {for which} FitNet succeeds, many videos contain {the} related sound of the ball. 
More analyses are given in Section \ref{sec:ablation} and the {\em supplementary}. 

As FitNet imposes a hard alignment by {regression} between {the} teacher and student, noisy side information from the audio teacher could bring a negative impact. 
Notably, our CCL is designed to close the cross-modal semantic gap and 
bring class labels {into} its loss formulation,  
thus showing robustness when distilling audio and (or) visual knowledge. 

\myparagraph{Video Retrieval.} 
In Table \ref{tab:retrieval}, 
we evaluate all the methods in the video retrieval task on {the} three setups of audio (A), visual (I) and audio-visual (AI) distillation on {the} UCF51 and ActivityNet datasets. This is to test the discriminability of the video representations in a challenging scenario that requires more fine-grained discrimination between videos. 

Our results in Table \ref{tab:retrieval} indicate that while the other alternative distillation methods do not always exhibit better performance compared to the baseline, our model outperforms the baseline consistently with large margins. For instance, the best competitor (CRD) performs well overall, but its performance of audio or visual distillation on UCF51 is not always stronger than the baseline. Specifically, R@1 in audio distillation (A), our CCL obtains an impressive 62.9 (vs 59.5 by CRD) on UCF51 although it stays behind CRD on ActivityNet (30.6 by ours vs 31.6 by CRD). On the other hand, in the case of R@1 in visual distillation (I), our CCL outperforms CRD with large margins on both datasets (66.8 vs 59.1 on UCF51 and 38.1 vs 32.3 on ActivityNet). 

Another observation is that, while many methods benefit from distilling audio-visual knowledge (AI), our model performs the best under this multi-modal setup. Our CCL significantly outperforms the state-of-the-art, giving a R@1 of 67.6 vs 64.3 by IFD on UCF51 and a R@1 of 39.5 vs 33.1 by CRD on ActivityNet. 
Using other metrics such as R@5 and R@10 on the smaller UCF51 dataset, the difference between {the} methods is smaller, e.g. on R@10 with audio-visual distillation, our CCL obtains 74.7 vs 74.2 by IFD. On the other hand, on the large-scale ActivityNet dataset, even on R@5 and R@10, our CCL significantly outperforms the state-of-the-art, e.g. on R@10 with audio-visual distillation, our CCL obtains 67.4 vs 61.7 by COR. These observations again suggest the robustness of our model for distilling knowledge from multiple modalities on different datasets, which are in line with the observations and trends on the video recognition task.

\begin{figure}[!t]
\centering
\includegraphics[width=0.475\textwidth]{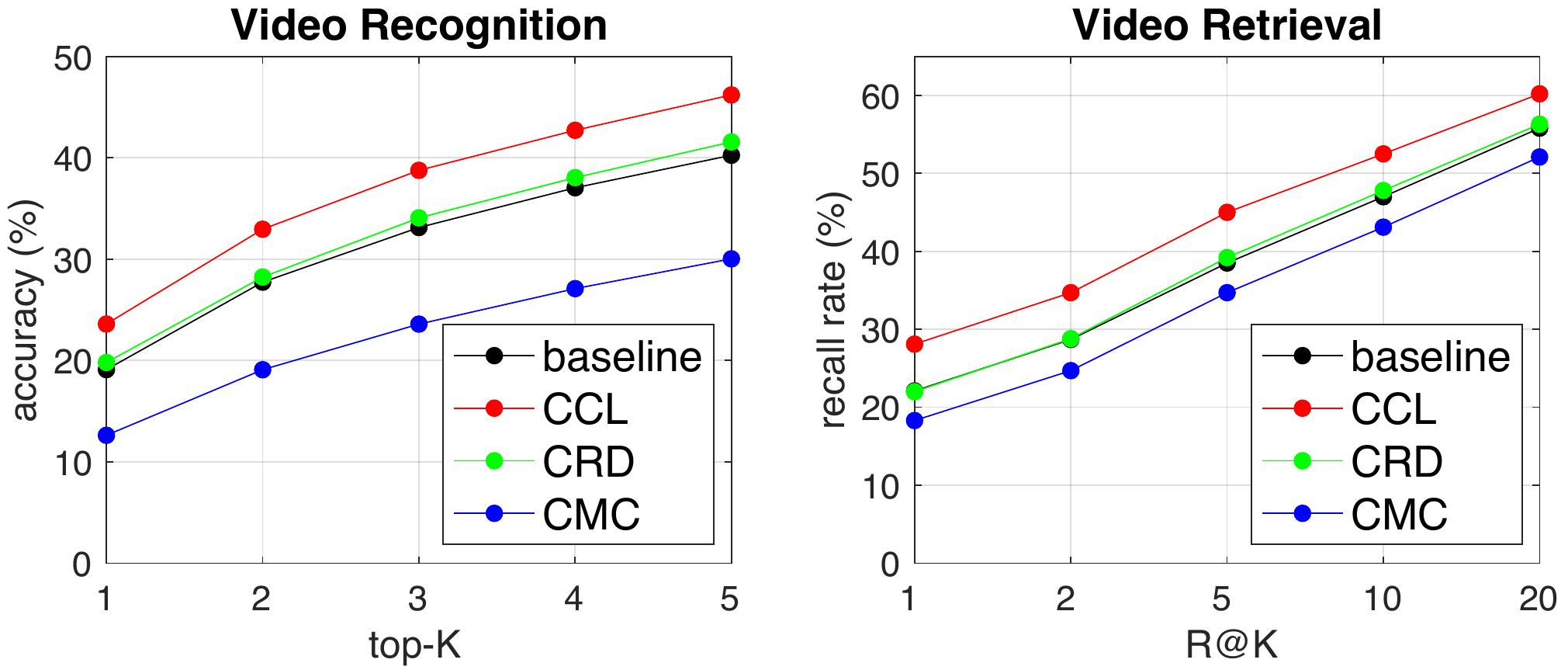}
\caption{
Results of CCL, CRD, CMC wrt the baseline on VGGSound. 
Left: recognition accuracy. 
Right: retrieval recall rate.  
}
\label{fig:vggsound}
\end{figure} 

\myparagraph{Experiments on the Large-scale VGGSound Dataset.} 
So far, CCL has shown its success in audio-visual distillation 
in human action and activity video datasets. 
To further test our model's generalisation on a challenging large-scale dataset of audio-visual events, we conduct experiments on the VGGSound dataset in the setup of audio-visual distillation (AI), where knowledge is jointly distilled from the audio and image modalities to improve the video modality. 
In addition to the baseline, we compare CCL to two strong competitors CRD, CMC that adopts contrastive learning. All methods are trained to predict the sound events in videos and tested for both video recognition and video retrieval tasks. 

As Figure \ref{fig:vggsound} shows, CCL (red curve) consistently outperforms the baseline (black curve) in both tasks, boosting the top-1 accuracy by 4.5\% and the R@1 by 6.0\% (exact numbers are given in {the} supplementary). 
While CRD (green curve) performs on par with the baseline, CMC performs lower than the baseline. 
Our results on VGGSound suggest the {ability of CCL to generalise} on a very large dataset, and demonstrate its capability to distill audio-visual knowledge for learning the video representations of sounds. Unlike the videos in UCF51 or ActivityNet, paired audio and video all share the same class semantics on VGGSound. This indicates the robustness of CCL in the different scenarios when the audio and video are either semantically correlated (VGGSound) or not always correlated (UCF51, ActivityNet). 

\subsection{Ablation Study and Qualitative Results}
\label{sec:ablation}

{
\renewcommand{\arraystretch}{1.1}
\begin{table}[!t]
    \centering
	\begin{tabular}{c|l|c c c}
        		& & \multicolumn{3}{c}{UCF51} \\
		& Method & A & I & AI \\ \hline 
		& baseline & 57.5 & 57.5 & 57.5 \\ \hline
		{\bf (a)} & CCL w/o composition & 63.2 & 65.8 & 66.9 \\ \hline 
		\multirow{3}{*}{\bf (b)} 
		& CCL w $\mathcal{L}_{ct}$ & 60.4 & 68.4 & 67.8 \\
		& CCL w/o $\mathcal{L}_{nce}$ & 63.1 & 67.4 & 66.3 \\ 
		& CCL w/o $\mathcal{L}_{JSD}$ & 64.0 & 67.8 & 68.2 \\ 
		\hline
		& CCL & \bf 64.9 & \bf 69.1 & \bf 70.0 
	\end{tabular}
 	\vskip 0.1em
	\caption{Ablation on {\bf (a)} model component: CCL w/o composition; 
	{\bf (b)} loss formulation: CCL w $\mathcal{L}_{ct}$; CCL w/o $\mathcal{L}_{nce}$; CCL w/o $\mathcal{L}_{JSD}$, 
	compared to the baseline and the CCL. 
	Metric: Top1 (\%).} 
	\label{tab:loss}
	\vskip -0.5em
\end{table}
}

To analyse our model formulation rationale, we conduct more studies on the UCF51 dataset in the following. 

\myparagraph{Ablation on Model Component.} 
In the model formulation,  
our main idea is to learn a compositional embedding 
that closes {the} cross-modal gap 
and captures task-relevant semantics. 
Rather than transferring knowledge across modalities directly, 
our CCL distills the unimodal knowledge from the teacher networks 
and the multi-modal knowledge from the composition functions collectively.  
To verify the idea of composition, we compare CCL to  
an ablative baseline (CCL w/o composition). 
Table \ref{tab:loss} {\bf(a)} shows that 
removing the composition 
degrades the recognition 
performance by 1.7\% (64.9-63.2), 3.3\% (69.1-65.8), 
3.1\% (70.0-66.9) on the setup of (A), (I), (AI). 
This supports our motivation to compose representations across modalities. 
As the compositional embedding is learned to rectify the 
teacher embedding as constrained by the task objective, 
it brings the task-relevant semantics to improve cross-modal distillation. 

\myparagraph{Ablation on Loss Formulation.} 
In {the} loss formulation, 
our goal is to associate positive pairs from the same class and 
disassociate negative ones. 
Our distillation objective brings class labels into contrastive learning, 
and performs alignment jointly in {the} {\em feature} and {\em prediction space} 
by the multi-class NCE ($\mathcal{L}_{nce}$) and the JSD loss ($\mathcal{L}_{JSD}$). 
To examine our objective empirically, 
we first compare our multi-class NCE $\mathcal{L}_{nce}$ (Eq. \eqref{eq:nce}) to  
an ablative baseline using the instance-level contrastive loss $\mathcal{L}_{ct}$ based on InfoNCE (Eq. \eqref{eq:infonce}). 
Table \ref{tab:loss} shows that CCL performs much better than the baseline ``CCL w $\mathcal{L}_{ct}$'', improving the accuracy by 
4.5\% (64.9-60.4), 2.2\% (70.0-67.8) on the setup of (A), (AI). 
This confirms the benefit of bringing class labels into contrastive learning. 
Next, we study the effect of similarity constraints 
in the {\em feature} and {\em prediction} space, 
we compare CCL to 
two ablative baselines: CCL w/o $\mathcal{L}_{nce}$, CCL w/o $\mathcal{L}_{JSD}$, 
which remove one constraint at a time. 
As Table \ref{tab:loss} shows, 
CCL performs the best. 
Removing $\mathcal{L}_{nce}$ 
decreases the performance of CCL by 
1.8\% (64.9-63.1), 1.7\% (69.1-67.4), 3.7\% (70.0-66.3) in the setup of (A), (I), (AI); 
while removing $\mathcal{L}_{JSD}$ 
also leads to performance degradation. 
These results indicate that $\mathcal{L}_{nce}$, $\mathcal{L}_{JSD}$ are complementary  
and work synergistically to distill knowledge across modalities.  

\begin{figure}[!t]
\centering
\includegraphics[width=0.46\textwidth]{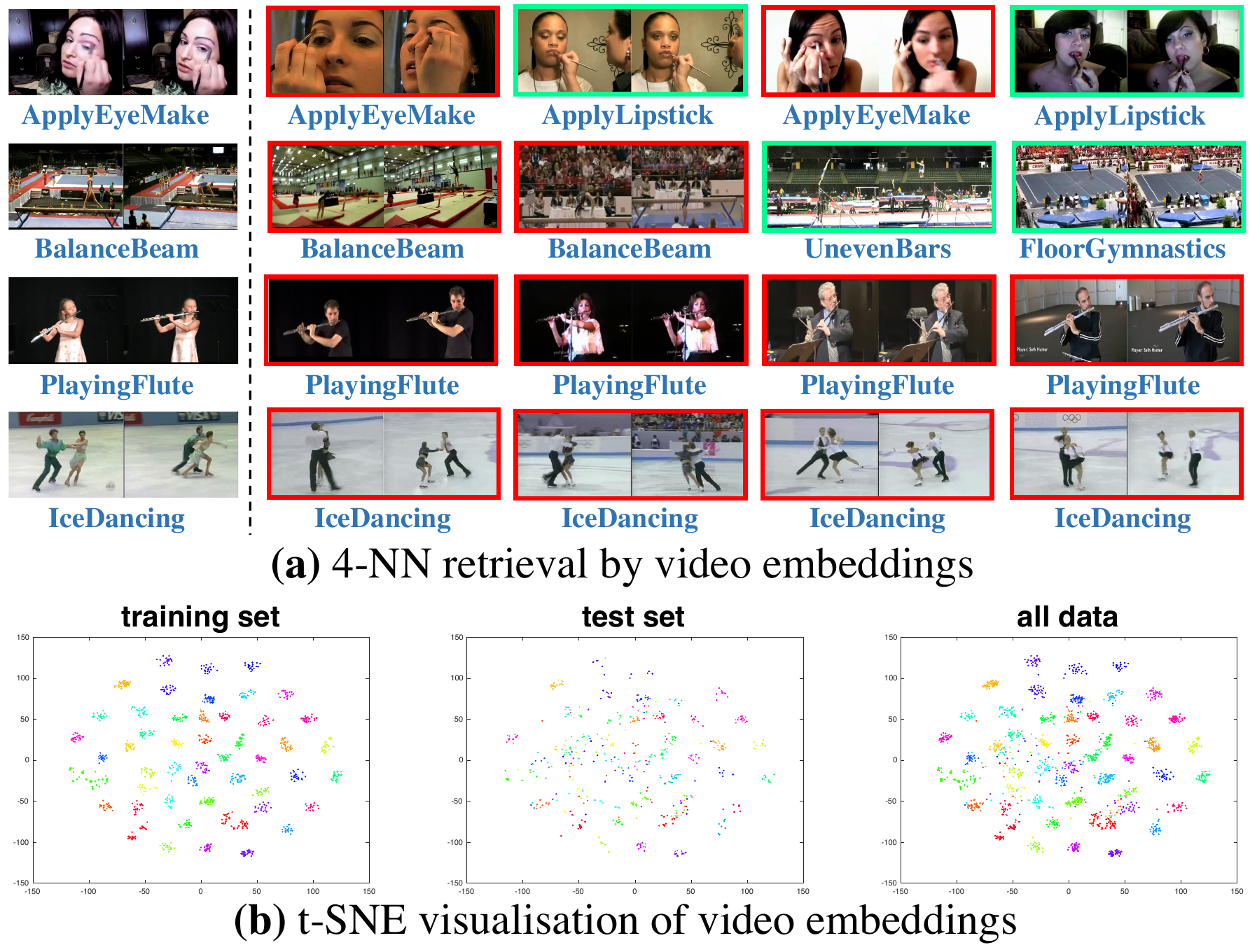}
\caption{Qualitative results {on UCF51}. 
{(a)} Left: query videos; Right: 4-NN retrieved items.  
{(b)} Visualisation with t-SNE \cite{maaten2008visualizing}. 
}
\label{fig:retrieval}
\vskip -0.3em
\end{figure}

\myparagraph{Qualitative Results.} 
To understand the video representations qualitatively, 
we analyse CCL with qualitative results. 
For k-NN retrieval (Figure \ref{fig:retrieval}{\bf (a)}), 
we observe that given the query videos, videos of the same or similar classes are retrieved,  
e.g. {for the} video ``{\em ice dancing}'', the top retrieved videos are from the same class. 
{For the} video ``{\em apply eye makeup}'', two videos are from the same class and the other two are from a similar action with subtle differences. 

When visualising the video embeddings (Figure \ref{fig:retrieval} {(b)}), 
we see that the embeddings of different classes (in different colours) 
are grouped into separated clusters; 
while the test set embeddings are lying on the manifolds 
similar to the training set. 
This means videos from training and test sets are grouped 
in a consistent way, where embeddings from the same class are associated with higher similarities. 
Our qualitative results overall show that CCL  
learns discriminative video representations from multi-modal distillation. 

\begin{figure}[!t]
\centering 
\includegraphics[width=0.475\textwidth]{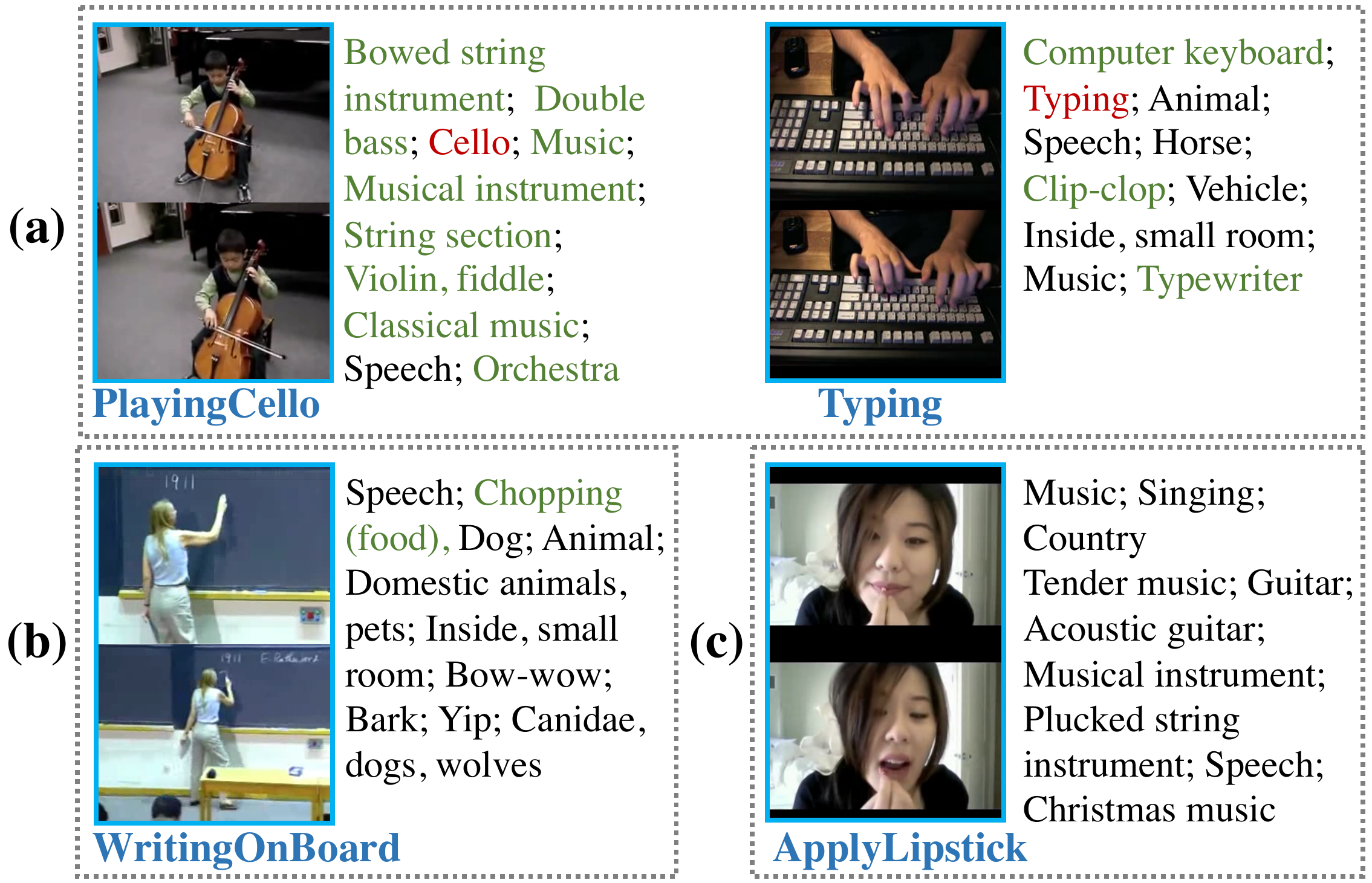}
\caption{
Audio-video correspondence: 
videos (labels in {\color{blue}blue}) 
from UCF51 and top-10 audio classes predicted by the audio network. 
Highly/weakly correlated audio events are in 
{\color{brickred}red}/{\color{darkgreen}green}. 
{(a)}, {(b)}, {(c)} denote video, audio are highly, weakly, or not correlated. 
}
\label{fig:audio}
\end{figure}

\myparagraph{Qualitative Analysis on Cross-Modal Correspondence.} 
To understand the cross-modal semantic gap, 
we provide visual examples of audio-video correspondence. 
As Figure \ref{fig:audio} shows, 
based on the top-10 predicted audio classes, 
we can manually distinguish 
the audio-video correspondence 
as highly, weakly, or not correlated. 
In (a), the audio event ``{\em cello}'' is highly related to the video action. 
In (b), the audio is dominated by speech but contains the sound ``{\em chopping}'' 
weakly related to the sound made by ``{\em writing on board}''. 
In (c), the audio is irrelevant ``{\em music}''. 
These evidences are in line with our assumption of {the} cross-modal semantic gap 
in unconstrained videos. 
Similarly, an image frame may not capture the whole video action, leading to {a} possible semantic gap between the image and video modalities. 
Notably, our model tackles this issue 
by introducing the compositional embeddings for compositional contrastive learning. 
More analyses are given in the {\em supplementary}. 

\section{Discussion and Conclusion}
We present a novel compositional contrastive learning (CCL) framework, 
a generic and effective approach to distill knowledge learned from heterogeneous data modalities for video representation learning. As there may exist {a} cross-modal semantic gap, we introduce the learnable compositional embeddings to close {the gap} and capture the task-related semantics. 
Our approach uniquely brings the unimodal knowledge (from teacher networks) and multi-modal knowledge (from composition functions) collectively to facilitate effective knowledge distillation. 
We compare our approach to a variety of state-of-the-art distillation methods, and demonstrate its performance advantages {for} both video recognition and video retrieval in different setups. Our empirical results also provide a realistic benchmark for future research in multi-modal distillation. As {a} future extension, our approach also opens up the possibility to bridge multiple modalities for multi-modal recognition and retrieval tasks. 

\vskip 0.5em 
\myparagraph{Acknowledgements} This work has been partially funded by the ERC (853489 - DEXIM) and by the DFG (2064/1 – Project number 390727645).

\newpage
{\small
\bibliographystyle{ieee_fullname}
\bibliography{reference}
}

\renewcommand\thesection{\Alph{section}}
\renewcommand\thefigure{\Alph{figure}}
\renewcommand\thealgorithm{\Alph{algorithm}}
\renewcommand\thetable{\Alph{table}}

\newpage

\begin{algorithm*}[!h]
	\caption{Compositional Contrastive Learning (Audio-Visual Distillation)} \label{Algorithm}
	\begin{algorithmic}[1]
	\REQUIRE 
	Video dataset $\mathcal{D} = \{\textbf{V}_i, y_i\}_{i=1}^{N}$, the corresponding image frames $\{{\textbf{I}_{ij}}\}_{j=1}^{M_i}$ and audio recording $\textbf{A}_i$ for each video. \\ 
	\REQUIRE 
	Trainable video student network $\theta_{\text{3D-CNN}}$. 
	Pre-trained audio and image teacher networks $\theta_{\text{1D-CNN}}, \theta_{\text{2D-CNN}}$. \\
	\FOR{$t=1$ \textbf{to} \textsl{max\_iter}} 
		\STATE
		$x_a \gets \theta_{\text{1D-CNN}}({\bf A}_i), 
		  x_i \gets \theta_{\text{2D-CNN}}({\bf I}_{ij}),
		  x_v \gets \theta_{\text{3D-CNN}}({\bf V}_i)$ 
		  \hfill\COMMENT{obtain unimodal audio, image, video embeddings} \\
		
		\STATE
		$x_{av} \gets \mathcal{F}_{av}(x_a, x_v) 
		  x_{iv} \gets \mathcal{F}_{iv}(x_i, x_v) $ 
		  \hfill\COMMENT{derive compositional embeddings} \\
		
		\STATE
		 $\mathcal{L}_{ce}^v \gets \mathcal{L}_{ce}(x_v,k), 
		   \mathcal{L}_{ce}^{av} \gets \mathcal{L}_{ce}(x_{av},k), 
		   \mathcal{L}_{ce}^{iv} \gets \mathcal{L}_{ce}(x_{iv},k)$ 
		   \hfill\COMMENT{compute video classification loss} \\
		
		\STATE
		$\mathcal{L}_{\text{audio}} \gets 
		  \lambda\mathcal{L}_{nce}(x_v, x_a){+}(1{-}\lambda)\mathcal{L}_{nce}(x_v, x_{av}){+}JSD(P_v||P_{av})$ 
		  \hfill\COMMENT{compute audio distillation loss} \\
		 
		 \STATE
		$\mathcal{L}_{\text{image}} \gets 
		  \lambda\mathcal{L}_{nce}(x_v, x_i){+}(1{-}\lambda)\mathcal{L}_{nce}(x_v, x_{iv}){+}JSD(P_v||P_{iv})$
		  \hfill\COMMENT{compute visual distillation loss} \\

		\STATE
		$\theta_{\text{3D-CNN}}^{t{+}1} \gets \theta_{\text{3D-CNN}}^{t}{-}\eta\frac{\partial \mathcal{L}^v}{\partial \theta_{\text{3D-CNN}}}$, where $\mathcal{L}^v{=}\mathcal{L}_{ce}^v{+}\mathcal{L}_{\text{audio}}{+}\mathcal{L}_{\text{image}}$ 
		\hfill\COMMENT{backprop on the video network} \\

		\STATE
		${\theta_{av}^{t+1}}\leftarrow{\theta_{av}^t}{-}\eta\frac{\partial \mathcal{L}_{ce}^{av}}{\partial \theta_{av}},
		{\theta_{iv}^{t+1}}\leftarrow{\theta_{iv}^t}{-}\eta\frac{\partial \mathcal{L}_{ce}^{iv}}{\partial \theta_{iv}}$ 
		\hfill\COMMENT{backprop on the composition functions} \\
	\ENDFOR
	\end{algorithmic}
\end{algorithm*}

\section{Additional Algorithm Details}

\begin{figure}[!h]
  \vskip -1em
  \centering
  \includegraphics[width=0.25\textwidth]{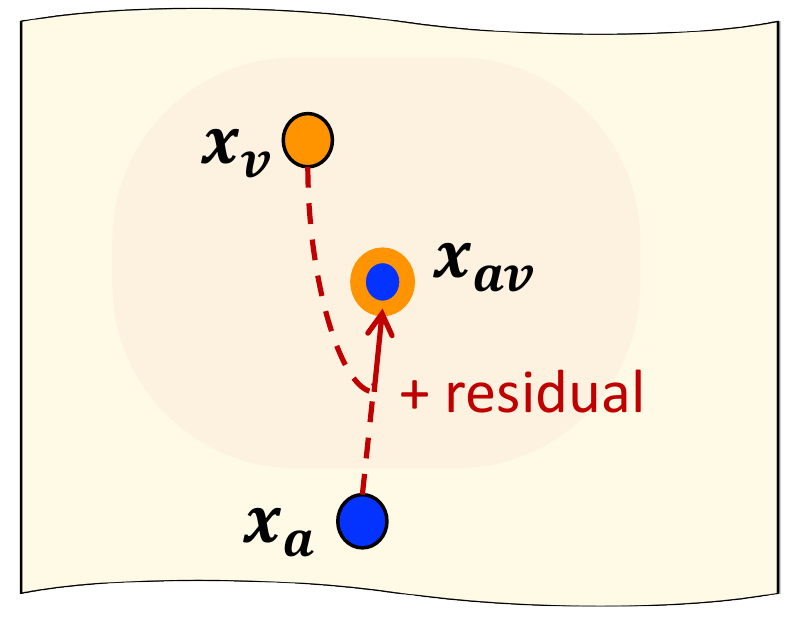}
   \caption{
     Illustration of the compositional embedding $x_{av}$. 
     The composition function uses residual learning to modify a teacher embedding $x_a$, 
     which shifts $x_a$ towards the video embedding $x_v$, resulting in
     the compositional embedding $x_{av}$. Given the classification constraint $\mathcal{L}_{ce}$,
     $x_{av}$ is enforced to share to the same video class as $x_v$, thus closing the possible cross-modal semantic gap.
     }
  \label{fig:composition}
\end{figure}

{Algorithm \ref{Algorithm}} gives an overview of our compositional contrastive learning (CCL) algorithm for audio-visual distillation. 
From an information-theoretic point of view, 
CCL distills audio-visual knowledge from the teacher networks 
by {maximising the mutual information} between the student network $\theta_{\text{3D-CNN}}$  
and 
the teacher networks $\theta_{\text{1D-CNN}}, \theta_{\text{2D-CNN}}$   
and the composition functions $\mathcal{F}_{av}, \mathcal{F}_{iv}$. 
While the multi-class contrastive loss $\mathcal{L}_{nce}$ 
contrasts the {feature similarity}, 
the Jensen–Shannon divergence $\mathcal{L}_{JSD}$ 
contrasts the {prediction similarity}, 
which together maximise the similarities between the cross-modal positive pairs from the same class. 
Importantly, {class labels} are introduced into both loss terms and the composition functions (Figure \ref{fig:composition}), 
thus ensuring to transfer the task-relevant knowledge to the student network. 

\section{Additional Analysis and Results}

\myparagraph{Analysis of Cross-Modal Correspondence.} 
Here, we first manually analyse the audio-video correspondence on UCF51. 
For each video, we compute the top-10 predicted audio classes using the audio network. For each video class, we compute the {top-10 associated audio classes} based on how frequently they are predicted as the top-10 audio classes. 
We summarise the video classes and their top-10 associated audio classes in Tables \ref{tab:ucf1}, \ref{tab:ucf2}, \ref{tab:ucf3},  
where we manually classify the audio-video correspondence as highly, weakly and not correlated, 
as summarised in Table \ref{tab:statistic}. 

{
\setlength{\tabcolsep}{2pt}
\renewcommand{\arraystretch}{1}
\begin{table}[!h]
\vskip -0.5em
    \centering
	\begin{tabular}{c|c|c}
	audio-video correspondence & \# video classes & proportion (\%) \\ \hline
	highly & 15 & 29.4 \\
	weakly & 15 & 29.4 \\
	not & 21 & 41.2 \\
	\end{tabular}
	\vskip 0.1em
	\caption{Statistics of the audio-video correspondence on UCF51. }
	\label{tab:statistic}
\end{table}
}
\begin{figure}[!t]
\centering
\includegraphics[width=0.48\textwidth]{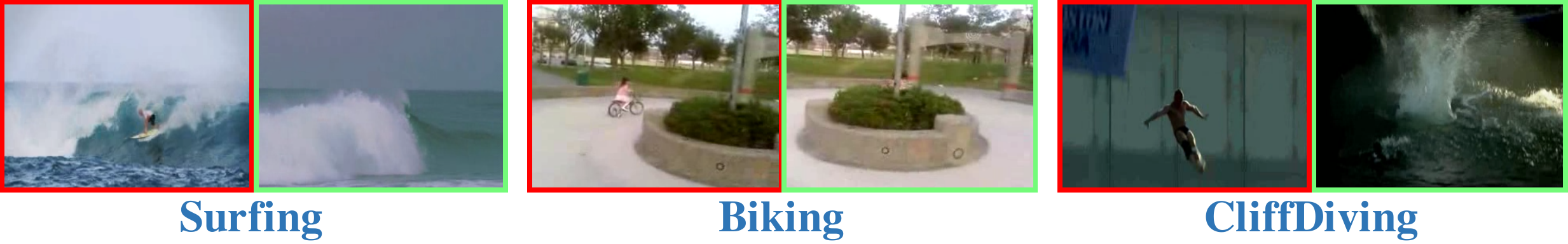}
\caption{
Image-video correspondence: 
videos (tagged in {\color{blue}blue}) from UCF. The 
{\color{brickred}red}/{\color{darkgreen}green} boxes mean that  
the visual cues in the image frames 
are {\color{brickred}highly}/{\color{darkgreen}weakly} correlated to the 
video classes. 
}
\label{fig:image}
\vskip -1em
\end{figure}
Moreover, we give some qualitative examples of the image-video correspondence (Figure \ref{fig:image}). As shown, the visual cues in image frames are generally highly or weakly related to the video content, e.g. the {\em sea} image is weakly correlated with the video class {\em skijet} due to occlusion.

\noindent 
{\em Remark.} Our analysis of the UCF51 dataset 
indicates the presence of a possible cross-modal semantic gap in multi-modal distillation. 
To empirically examine how CCL deal with this issue in practice, 
we evaluate CCL and its best competitor CRD using highly/weakly or not correlated audios for audio distillation. 
As Table \ref{tab:results} shows, on the 21 classes with not correlated audios, CRD performs on par with the baseline w/o distillation (+0.5\% acc), while our CCL outperforms the baseline significantly (+3.1\% acc). 
This suggests that our CCL can distill complementary information from audio even if it is uncorrelated with the video.

{
\setlength{\tabcolsep}{2pt}
\renewcommand{\arraystretch}{1.2}
 \begin{table}[!h]
   \vskip -0.5em
     \centering
 	\begin{tabular}{c|c|c|c}
 	audio-video correspondence & baseline & CRD & \bf CCL \\ \hline
  	weak/highly {correlated} (30 classes) & 55.3 & 59.6 & \bf 65.5 \\
 	not {correlated} (21 classes) & 61.0 & 61.5 & \bf 64.1 \\
 	\end{tabular}
	\vskip 0.1em
 	\caption{Ablation study of audio distillation on UCF51. }
 	\label{tab:results}
 \end{table}
 }
 
\myparagraph{Tabular Results on VGGSound.}
In Table \ref{tab:vggsound}, we provide the tabular results of audio-visual distillation on the large-scale VGGSound dataset, which includes three contrastive learning methods (CRD, CMC, CCL) in comparison to the baseline for the video recognition and video retrieval tasks. 

{
\setlength{\tabcolsep}{7pt}
\renewcommand{\arraystretch}{1.2}
\begin{table}[!h]
\vskip -0.5em
	\small
	\centering
	\resizebox{\columnwidth}{!}{
	\begin{tabular}{l|cc|cccc}	
	    	Task 
		& \multicolumn{2}{c|}{Recognition} 
		& \multicolumn{4}{c}{Retrieval} \\ \cline{2-7}
		Metric
		& Top1 & Top5 
		& R1 & R5 & R10 & R20 \\ \hline
		baseline & 
		19.1 & 20.3 &
		22.1 & 38.5 & 47.0 & 55.8
		\\ \hline
		CRD & 
		19.8 & 41.6 & 
		22.0 & 39.2 & 47.8 & 56.3 \\
		CMC & 
		12.6 & 30.0 & 
		18.3 & 34.7 & 43.1 & 52.1\\
		\hline 
		\bf CCL & 
		\bf 23.6 & \bf 46.2 &
		\bf 28.1 & \bf 45.0 & \bf 52.5 & \bf 60.2 \\
	\end{tabular}
	}
        \vskip 0.1em
	\caption{Audio-visual distillation on VGGSound.} 
	\label{tab:vggsound}
\end{table}
}

{
\setlength{\tabcolsep}{5pt}
\renewcommand{\arraystretch}{1.15}
\begin{table*}[!h]
\small
	\centering
	\begin{tabular}{l|l|l}
	\hlineB{2}
	\bf Video Class & \bf Top-10 Associated Audio Classes & \bf Correlated \\ \hline
    \multirow{2}{75pt}{ApplyEyeMakeup} 
    & Speech; Inside, small room; Music; Female speech, woman speaking; 
    & \multirow{2}{30pt}{not} \\
    & Vehicle; Writing; Conversation; Narration, monologue; Animal; Rustle \\ \hline
    
    \multirow{2}{75pt}{ApplyLipstick}
     & Speech; Music; Inside, small room; Vehicle; Animal; Female speech, 
     & \multirow{2}{30pt}{not} \\
     & woman speaking; Narration, monologue; Conversation; Musical instrument; Writing \\ \hline
	
	\multirow{2}{75pt}{Archery}
	 & Speech; Music; Vehicle; {\color{brickred}Arrow}; Inside, small room; Outside, 
	 & \multirow{2}{30pt}{highly} \\
	 & rural or natural; Animal; Car; Door; Bird \\ \hline
	 
	 \multirow{2}{75pt}{BabyCrawling}
	 & Speech; Inside, small room; Music; Animal; {\color{darkgreen}Child speech, kid speaking}; {\color{darkgreen}Babbling}; 
	 & \multirow{2}{30pt}{weakly} \\
	 & Laughter; Domestic animals, pets; Vehicle; Crying, sobbing \\ \hline
	 
	 \multirow{2}{75pt}{BalanceBeam}
	 & Speech; Music; Vehicle; Outside, urban or manmade; Crowd; Inside, 
	 & \multirow{2}{30pt}{weakly} \\
	 & large room or hall; Inside, public space; {\color{darkgreen}Basketball bounce}; Car; Animal \\ \hline
	 
	 \multirow{2}{75pt}{BandMarching}
	 & {\color{brickred}Music}; Speech; {\color{brickred}Musical instrument}; {\color{brickred}Drum}; {\color{brickred}Percussion}; Crowd; {\color{brickred}Orchestra}; 
	 & \multirow{2}{30pt}{highly} \\
	 & {\color{brickred}Brass instrument}; Vehicle; Wood block \\ \hline
	 
	 \multirow{2}{75pt}{BasketballDunk}
	 & Music; Speech; Vehicle; {\color{brickred}Basketball bounce}; Outside, urban or manmade; 
	 & \multirow{2}{30pt}{highly} \\
	 & Crowd; Car; Hip hop music; Slam; Singing \\ \hline
	 
	 \multirow{2}{75pt}{BlowDryHair}
	 & Music; Speech; Vehicle; Inside, small room; {\color{brickred} Hair dryer}; Vacuum cleaner; 
	 & \multirow{2}{30pt}{highly} \\
	 & Car; Mechanical fan; Animal; Train \\ \hline
	 
	 \multirow{2}{75pt}{BlowingCandles}
	 & Speech; Inside, small room; Music; Animal; Laughter; Chuckle, chortle; 
	 & \multirow{2}{30pt}{not} \\
	 & Snicker; Child speech, kid speaking; Inside, large room or hall; Domestic animals, pets \\ \hline
	 
	 \multirow{2}{75pt}{BodyWeightSquats}
	 & Speech; Music; Vehicle; Inside, small room; Male speech, man speaking; 
	 & \multirow{2}{30pt}{not} \\
	 & Narration, monologue; Animal; Musical instrument; Car; Conversation \\ \hline
	 
	 \multirow{2}{75pt}{Bowling}  
	 & Speech; Music; Vehicle; Outside, urban or manmade; Train; 
	 & \multirow{2}{30pt}{weakly} \\
	 & {\color{darkgreen}Slam}; Car; Animal; Inside, public space; Inside, large room or hall \\ \hline
	 
	 \multirow{2}{75pt}{BoxingPunchingBag} 
	 & Speech; Music; {\color{darkgreen}Slam}; Inside, large room or hall; Inside, small room; 
	 & \multirow{2}{30pt}{weakly} \\ 
	 & {\color{darkgreen}Thump, thud}; Singing; Tap; Musical instrument; Vehicle \\ \hline
	
	\multirow{2}{75pt}{BoxingSpeedBag} 
	& Speech; Vehicle; Engine; Music; Car; {\color{darkgreen}Idling}; Machine gun; 
	& \multirow{2}{30pt}{weakly} \\ 
	& Outside, urban or manmade; Engine starting; Motorcycle \\ \hline
	
	\multirow{2}{75pt}{BrushingTeeth} 
	 & Speech; Inside, small room; Music; Animal; {\color{brickred} Toothbrush}; Domestic animals, 
	 & \multirow{2}{30pt}{highly} \\
	 & pets; Scratch; Rub; Vehicle; Child speech, kid speaking \\ \hline
	
	\multirow{2}{75pt}{CliffDiving} 
	 & Music; Speech; Vehicle; Musical instrument; Electronic music; 
	 & \multirow{2}{30pt}{not} \\
	 & Car; Rock music; Outside, urban or manmade; Guitar; Trance music \\ \hline
	
	\multirow{2}{75pt}{CricketBowling} 
	 & Speech; Music; Vehicle; Outside, urban or manmade; Outside, rural or natural; 
	 & \multirow{2}{30pt}{weakly} \\
	 & Car; Animal; {\color{darkgreen}Basketball bounce}; {\color{darkgreen}Slam}; Inside, large room or hall \\ \hline
	 
	 \multirow{2}{75pt}{CricketShot} 
	 & Speech; Music; Vehicle; Outside, urban or manmade; Arrow; Animal; 
	 & \multirow{2}{30pt}{weakly} \\
	 & Outside, rural or natural; {\color{darkgreen}Slam}; Car; {\color{darkgreen}Thump, thud} \\ 
	 
	\hlineB{2} 
	\end{tabular}
	\vskip 0.5em
	\caption{
	Audio-video correspondence on the UCF51 classes.
	Video classes (1˜17), the top-10 associated audio classes, 
	and the audio-video correlation: highly, weakly, or not correlated. 
	Note: audio events highly/weakly correlated with the video are highlighted 
	in {\color{brickred}red}/{\color{darkgreen}green}.
	} 
	\label{tab:ucf1}
\end{table*}
}

{
\setlength{\tabcolsep}{5pt}
\renewcommand{\arraystretch}{1.15}
\begin{table*}[!h]
\small
	\centering
	\begin{tabular}{l|l|l}
	\hlineB{2}
	\bf Video Class & \bf Top-10 Associated Audio Classes & \bf Correlated \\ \hline

	 \multirow{2}{75pt}{CuttingInKitchen} 
	 & Music; Speech; Inside, small room; {\color{brickred} Chopping (food)}; {\color{darkgreen}Dishes, pots, and pans}; 
	 & \multirow{2}{30pt}{highly} \\
	 & Wood; Animal; {\color{brickred} Chop}; Vehicle; {\color{darkgreen}Cutlery, silverware} \\ \hline
	 
	 \multirow{2}{75pt}{FieldHockeyPenalty} 
	 & Speech; Vehicle; Music; Outside, urban or manmade; {\color{darkgreen}Basketball bounce}; 
	 & \multirow{2}{30pt}{weakly} \\
	 & Car; Animal; Crowd; Outside, rural or natural; Hubbub, speech noise, speech babble \\ \hline
	 
	 \multirow{2}{75pt}{FloorGymnastics} 
	 & Music; Speech; Crowd; Cheering; Vehicle; Inside, large room or hall; 
	 & \multirow{2}{30pt}{not} \\
	 & Children shouting; Outside, urban or manmade; Singing; Whoop \\ \hline

     \multirow{2}{75pt}{FrisbeeCatch} 
	 & Speech; Music; Vehicle; Outside, urban or manmade; Singing; 
	 & \multirow{2}{30pt}{not} \\
	 & Car; Pop music; Hubbub, speech noise, speech babble; Boat, Water vehicle; Crowd \\ \hline
	 
	 \multirow{2}{75pt}{FrontCrawl} 
	 & Speech; Vehicle; Music; {\color{darkgreen}Water}; Stream; Car; Boat, Water vehicle; 
	 & \multirow{2}{30pt}{weakly} \\
	 & Outside, urban or manmade; Splash, splatter; Outside, rural or natural\\ \hline
	 
    \multirow{2}{75pt}{Haircut} 
    & Speech; Music; Inside, small room; Vehicle; Musical instrument; Animal; Inside, large room or hall; 
    & \multirow{2}{30pt}{not} \\
    & Electronic music; Outside, urban or manmade; Female speech, woman speaking \\ \hline
    
    \multirow{2}{75pt}{HammerThrow} 
    & Music; Speech; Vehicle; Outside, urban or manmade; Car; Male speech, 
    & \multirow{2}{30pt}{not} \\
    & man speaking; Animal; Musical instrument; Outside, rural or natural; Basketball bounce \\ \hline
    
    \multirow{2}{75pt}{Hammering} 
    & Music; Speech; {\color{brickred} Hammer}; Whack, thwack; Chop; Tools; 
    & \multirow{2}{30pt}{highly} \\
    & Inside, small room; Musical instrument; Vehicle; Glass \\ \hline
    
    \multirow{2}{75pt}{HandstandPushups} 
    & Speech; Music; Vehicle; Animal; Inside, small room; Musical instrument; 
    & \multirow{2}{30pt}{not} \\
    & Singing; Domestic animals, pets; Car; Pink noise \\ \hline
    
    \multirow{2}{75pt}{HandstandWalking} 
    & Speech; Music; Vehicle; Animal; Outside, urban or manmade; Car; 
    & \multirow{2}{30pt}{not} \\
    & Inside, small room; Musical instrument; Singing; Domestic animals, pets \\ \hline
    
    \multirow{2}{75pt}{HeadMassage} 
    & Speech; Music; Vehicle; Outside, urban or manmade; Inside, small room; 
    & \multirow{2}{30pt}{not} \\
    & Musical instrument; Singing; Animal; Inside, large room or hall; Car \\ \hline
    
    \multirow{2}{75pt}{IceDancing} 
    & Music; Speech; Musical instrument; Vehicle; Orchestra; Singing; 
    & \multirow{2}{30pt}{not} \\
    & Theme music; Outside, urban or manmade; Television; Guitar \\ \hline
    
    \multirow{2}{75pt}{Knitting} 
    & Speech; Inside, small room; Music; Writing; Animal; Vehicle; 
    & \multirow{2}{30pt}{not} \\
    & Female speech, woman speaking; Conversation; Air conditioning; Narration, monologue \\ \hline
    
    \multirow{2}{75pt}{LongJump} 
    & Speech; Music; Outside, urban or manmade; Vehicle; Car; Basketball bounce; 
    & \multirow{2}{30pt}{weakly} \\
    & Inside, public space; Crowd; Hubbub, speech noise, speech babble; {\color{darkgreen}Run} \\ \hline
    
    \multirow{2}{75pt}{MoppingFloor} 
    & Speech; Inside, small room; Music; Animal; Vehicle; Inside, large room or hall; 
    & \multirow{2}{30pt}{not} \\
    & Male speech, man speaking; Television; Narration, monologue; Domestic animals, pets \\ \hline
    
    \multirow{2}{75pt}{ParallelBars} 
    & Speech; Music; Crowd; Inside, large room or hall; Inside, public space; 
    & \multirow{2}{30pt}{weakly} \\
    & Outside, urban or manmade; {\color{darkgreen}Basketball bounce}; Cheering; Slam; Vehicle \\ \hline
    
    \multirow{2}{75pt}{PlayingCello} 
    & {\color{darkgreen}Music}; {\color{darkgreen}Musical instrument}; {\color{darkgreen}Bowed string instrument}; {\color{brickred}Cello}; {\color{darkgreen}String section}; 
    & \multirow{2}{30pt}{highly} \\
    & Violin, fiddle; Double bass; {\color{darkgreen}Classical music}; Orchestra; Piano \\ 
    
	\hlineB{2} 
	\end{tabular}
	\vskip 0.5em
	\caption{
	Audio-video correspondence on the UCF51 classes.
	Video classes (18˜34), the top-10 associated audio classes, 
	and the audio-video correlation: highly, weakly, or not correlated. 
	Note: audio events highly/weakly correlated with the video are highlighted 
	in {\color{brickred}red}/{\color{darkgreen}green}.
	} 
	\label{tab:ucf2}
\end{table*}
}

{
\setlength{\tabcolsep}{5pt}
\renewcommand{\arraystretch}{1.15}
\begin{table*}[!h]
\small
	\centering
	\begin{tabular}{l|l|l}
	\hlineB{2}
	\bf Video Class & \bf Top-10 Associated Audio Classes & \bf Correlated \\ \hline

    \multirow{2}{75pt}{PlayingDaf} 
    & {\color{darkgreen}Music}; {\color{brickred}Drum}; {\color{darkgreen}Musical instrument}; {\color{brickred}Percussion}; {\color{darkgreen}Drum kit}; 
    & \multirow{2}{30pt}{highly} \\
    & {\color{darkgreen}Bass drum}; {\color{darkgreen}Snare drum}; {\color{darkgreen}Drum roll}; Wood block; {\color{brickred}Rimshot} \\ \hline
    
    \multirow{2}{75pt}{PlayingDhol} 
    & {\color{darkgreen}Music}; {\color{brickred}Drum}; {\color{darkgreen}Musical instrument}; {\color{brickred}Percussion}; {\color{darkgreen}Drum kit}; {\color{darkgreen}Bass drum}; 
    & \multirow{2}{30pt}{highly} \\
    & Wood block; Speech; {\color{darkgreen}Snare drum}; {\color{darkgreen}Tabla} \\ \hline
    
    \multirow{2}{75pt}{PlayingFlute} 
    & {\color{darkgreen}Musical instrument}; {\color{darkgreen}Music}; {\color{brickred} Flute}; {\color{darkgreen}Wind instrument, woodwind instrument}; {\color{darkgreen}Classical music}; 
    & \multirow{2}{30pt}{highly} \\
    & Inside, small room; {\color{darkgreen}Bowed string instrument}; Piano; Violin, fiddle; Speech \\ \hline
    
    \multirow{2}{75pt}{PlayingSitar} 
    & {\color{darkgreen}Music}; {\color{darkgreen}Musical instrument}; {\color{brickred}Sitar}; {\color{darkgreen}Plucked string instrument}; 
    & \multirow{2}{30pt}{highly} \\
    & {\color{darkgreen}Classical music}; {\color{darkgreen}Carnatic music}; {\color{darkgreen}Bowed string instrument}; Speech; Tabla; {\color{darkgreen}Music of Asia} \\ \hline
    
    \multirow{2}{75pt}{Rafting} 
    & Music; Speech; Vehicle; Singing; Musical instrument; 
    & \multirow{2}{30pt}{weakly} \\
    & {\color{darkgreen}Waves, surf}; Ocean; Car; {\color{darkgreen}Waterfall}; Guitar \\ \hline

    \multirow{2}{75pt}{ShavingBeard}   
    & Speech; Inside, small room; Music; Vehicle; Animal; Inside, large room or hall; 
    & \multirow{2}{30pt}{highly} \\
    & {\color{brickred} Electric shaver, electric razor}; Outside, urban or manmade; Electric toothbrush; Buzz \\ \hline
    
    \multirow{2}{75pt}{Shotput}   
    & Speech; Music; Vehicle; Outside, urban or manmade; Animal; Car; 
    & \multirow{2}{30pt}{not} \\
    & Outside, rural or natural; Musical instrument; Hubbub, speech noise, speech babble; Singing \\ \hline
    
    \multirow{2}{75pt}{SkyDiving}  
    & Music; Musical instrument; Singing; Punk rock; Rock music; Heavy metal; 
    & \multirow{2}{30pt}{not} \\
    & Grunge; Progressive rock; Angry music; Rock and roll \\ \hline
    
    \multirow{2}{75pt}{SoccerPenalty} 
    & Speech; Outside, urban or manmade; Music; Vehicle; {\color{darkgreen}Basketball bounce}; 
    & \multirow{2}{30pt}{weakly} \\
    & Crowd; Slam; Male speech, man speaking; Car; Inside, public space \\ \hline
    
    \multirow{2}{75pt}{StillRings} 
    & Speech; Music; Vehicle; Outside, urban or manmade; Car; Inside, public space; 
    & \multirow{2}{30pt}{not} \\
    & Basketball bounce; Crowd; Inside, large room or hall; Slam \\ \hline
    
    \multirow{2}{75pt}{SumoWrestling} 
    & Speech; Music; Crowd; Inside, large room or hall; Inside, public space; 
    & \multirow{2}{30pt}{weakly} \\
    & Outside, urban or manmade; Cheering; Chatter; Basketball bounce; {\color{darkgreen}Slam} \\ \hline
    
    \multirow{2}{75pt}{Surfing} 
    & Music; Musical instrument; Vehicle; Speech; Rock music; 
    & \multirow{2}{30pt}{not} \\
    & Rock and roll; Punk rock; Guitar; Singing; Car \\ \hline

    \multirow{2}{75pt}{TableTennisShot} 
    & Speech; Music; {\color{brickred}Ping}; Animal; Tap; Inside, small room; 
    & \multirow{2}{30pt}{highly} \\
    & Inside, large room or hall; Vehicle; Whack, thwack; Bouncing \\ \hline

    \multirow{2}{75pt}{Typing} 
    & {\color{brickred}Typing}; Speech; {\color{darkgreen}Computer keyboard}; {\color{darkgreen}Typewriter}; Vehicle;
    & \multirow{2}{30pt}{highly} \\
    & Inside, small room; Music; Animal; Engine; Sewing machine \\ \hline
    
    \multirow{2}{75pt}{UnevenBars} 
    & Music; Speech; Outside, urban or manmade; Vehicle; Crowd; Basketball bounce; 
    & \multirow{2}{30pt}{not} \\
    & Car; Cheering; Slam; Children shouting \\ \hline
    
    \multirow{2}{75pt}{WallPushups} 
    & Speech; Music; Inside, small room; Animal; Vehicle; Narration, monologue; 
    & \multirow{2}{30pt}{not} \\
    & Male speech, man speaking; Conversation; Female speech, woman speaking; Musical instrument \\ \hline
    
    \multirow{2}{75pt}{WritingOnBoard} 
    & Speech; Inside, small room; Music; Male speech, man speaking; Narration, monologue; 
    & \multirow{2}{30pt}{weakly} \\
    & Inside, large room or hall; {\color{darkgreen}Chopping (food)}; {\color{darkgreen}Writing}; Chop; Conversation \\ \hline

	\hlineB{2} 
	\end{tabular}
	\vskip 0.5em
	\caption{
	Audio-video correspondence on the UCF51 classes.
	Video classes (35˜51), the top-10 associated audio classes, 
	and the audio-video correlation: highly, weakly, or not correlated. 
	Note: audio events highly/weakly correlated with the video are highlighted 
	in {\color{brickred}red}/{\color{darkgreen}green}.
	} 
	\label{tab:ucf3}
\end{table*}
}

\end{document}